\documentclass[letterpaper, 10 pt, journal, twoside]{IEEEtran}
\usepackage{epsfig}
\usepackage{amssymb}
\usepackage{cite}
\usepackage{booktabs}
\usepackage{multirow}
\usepackage{subcaption}
\usepackage{mathtools}
\usepackage{siunitx}
\usepackage{comment}
\usepackage{threeparttable}
\usepackage{bm}
\usepackage{color}

\usepackage{amssymb}% http://ctan.org/pkg/amssymb
\usepackage{pifont}% http://ctan.org/pkg/pifont
\ifCLASSINFOpdf
\else
\fi
\usepackage{amsmath}
\usepackage{algorithmic}
\usepackage{array}
\newcommand{\eg}{\textit{e}.\textit{g}.,}
\begin{document}
\title{\LARGE \bf Toward an Affective Touch Robot: Subjective and Physiological Evaluation of Gentle Stroke Motion Using a Human-Imitation Hand}

% \author{Michael~Shell,~\IEEEmembership{Member,~IEEE,}
%         John~Doe,~\IEEEmembership{Fellow,~OSA,}
%         and~Jane~Doe,~\IEEEmembership{Life~Fellow,~IEEE}% <-this % stops a space
% \thanks{M. Shell was with the Department
% of Electrical and Computer Engineering, Georgia Institute of Technology, Atlanta,
% GA, 30332 USA e-mail: (see http://www.michaelshell.org/contact.html).}% <-this % stops a space
% \thanks{J. Doe and J. Doe are with Anonymous University.}% <-this % stops a space
% \thanks{Manuscript received April 19, 2005; revised August 26, 2015.}}
\author{Tomoki Ishikura$^{1}$, Akishige Yuguchi$^{1}$, Yuki Kitamura$^{1}$, Sung-Gwi Cho$^{1}$, Ming Ding$^{1,2}$, 
%Gustavo Alfonso {Garcia Ricardez}$^{1}$,\\
Jun Takamatsu$^{1}$,\\ Wataru Sato$^{3}$, Sakiko Yoshikawa$^{4}$, and Tsukasa Ogasawara$^{1}$%
\thanks{$^{1}$T. Ishikura, A. Yuguchi, Y. Kitamura, S.-G. Cho, M. Ding, %G. A. Garcia Ricardez,
J. Takamatsu, and T. Ogasawara are with Division of Information Science, 
        Nara Institute of Science and Technology, 8916-5, Takayama, Ikoma, Nara, 630-0192, Japan.
        {\tt\small \{ishikura.tomoki.iq0, %kitamura.yuki.ks3, 
        yuguchi.akishige.xu9, cho.sungi.cg3, ding, j-taka, ogasawar\}@is.naist.jp}}%
\thanks{$^{2}$M. Ding is also with %Tier IV Intelligent Vehicle Design and Development Center, 
Institutes of Innovation for Future Society, Nagoya University, Furo-cho, Chikusa-ku, Nagoya 464-8601, Japan.}
\thanks{$^{3}$W. Sato is with Psychological Process Team, BZP, Robotics Project, RIKEN, 2-2-2 Hikaridai, Seika-cho, Soraku-gun, Kyoto 619-0288, Japan.
        {\tt\small \{wataru.sato.ya\}@riken.jp}}
\thanks{$^{4}$S. Yoshikawa is with Faculty of Art and Design, Kyoto University of the Arts, 2-116 Kitashirakawauryuzan-cho, Sakyo, Kyoto, 606-8271, Japan.
        {\tt\small \{yoshikawa.sakiko.4n\}@kyoto-u.ac.jp}}
}
\maketitle

\begin{abstract}
% Affective touch affords positive effects such as mitigation of stress and pain. 
Affective touch offers positive psychological and physiological benefits such as the mitigation of stress and pain. 
% If a robot could realize human-like affective touch, the robot opens the new application area, such as supporting care work.
If a robot could realize human-like affective touch, it would open up new application areas, including supporting care work. 
% In this research, toward an affective touch robot, we focus on gentle stroking by a robot to evoke the same emotions as stroking by a human.
In this research, we focused on the gentle stroking motion of a robot to evoke the same emotions that human touch would evoke: in other words, an affective touch robot. 
% We propose a robot system to gently stroke the back of a human using our designed human-imitation hand. 
We propose a robot that is able to gently stroke the back of a human using our designed human-imitation hand. 
% To evaluate the emotional effects, we compare the combination of two agents (human-imitation hand and human hand) and two stroke speeds (3 cm/s and 30 cm/s). 
To evaluate the emotional effects of this affective touch, we compared the results of a combination of two agents (the human-imitation hand and the human hand), at two stroke speeds (3 and 30 cm/s). 
% The result of subjective and physiological evaluations showed the following three findings: 1) the subjects evaluate strokes at a similar tendency in the speed aspect with both human hand and human-imitation hand in both subjective and physiological evaluations; 2) the subjects feel more pleasant and arousal with the faster-speed (30 cm/s than 3 cm/s) stroke; 3) less fitting due to bending a back gets the bad emotional effect.
The results of the subjective and physiological evaluations highlighted the following three findings: 1) the subjects evaluated strokes similarly with regard to the stroke speed of the human and human-imitation hand, in both the subjective and physiological evaluations; 2) the subjects felt greater pleasure and arousal at the faster stroke rate (30 cm/s rather than 3 cm/s); and 3) poorer fitting of the human-imitation hand due to the bending of the back had a negative emotional effect on the subjects.
\end{abstract}

% \begin{IEEEkeywords}
% Robotic Assembly, Jamming Gripper, Soft Robotics
% \end{IEEEkeywords}

% \fi
\IEEEpeerreviewmaketitle
\section{Introduction}
\label{intro}
Affective touch provides us with positive psychological effects.
For example, Henricson~\textit{et~al.}~\cite{henricson2008outcome} showed that affective touch stabilizes patients’ mental and circulatory dynamics.
Anderson~\textit{et~al.}~\cite{andersson2009tactile} showed that affective touch reduces patients’ anxiety, stress, and pain, and as a result, improves the quality of sleep.
Suzuki~\textit{et~al.}~\cite{suzuki2010physical} showed that affective touch reduces the chance of aggression in elderly dementia patients. In short, if a robot could realize an affective touch, it would be supportive of care work.

% There are many research methods to realize affective touch by a robot. 
There have been many research approaches to accomplish affective touch by a robot. 
% Toyoshima~\textit{et~al.}~\cite{toyoshima2018WELCARO} showed that a robot hand to gently stroke humans should have the mechanism to fit the stroking surface and the warm temperature.
Toyoshima~\textit{et~al.}~\cite{toyoshima2018WELCARO} showed that a robot hand used to gently stroke humans should have a mechanism that fits the stroking surface, and operate at a warm temperature.
% Koizumi~\textit{et~al.}~\cite{koizumi2017} mentioned that in gentle stroke, keeping pressure and velocity achieves positive emotional effects.
Koizumi~\textit{et~al.}~\cite{koizumi2017} mentioned that maintaining pressure and velocity achieved positive emotional effects when stroking gently.
% However, there still remain many unknown factors for affective touch by a robot, such as rigidity of the hand and actual velocity of stroking.
However, there are still many unknown factors regarding affective touch by a robot, such as the rigidity of the hand and the actual velocity of the stroking motion.
%%%%%%%%%%%%%%%%%%%%%%%%%%%%%%%%%%%%%%%%%%%%%%%%
\begin{figure}[t]
  \begin{center}
    \begin{tabular}{c}
      % 1
      \begin{minipage}{0.45\hsize}
        \begin{center}
          \includegraphics[keepaspectratio=true, width=1.0\columnwidth]{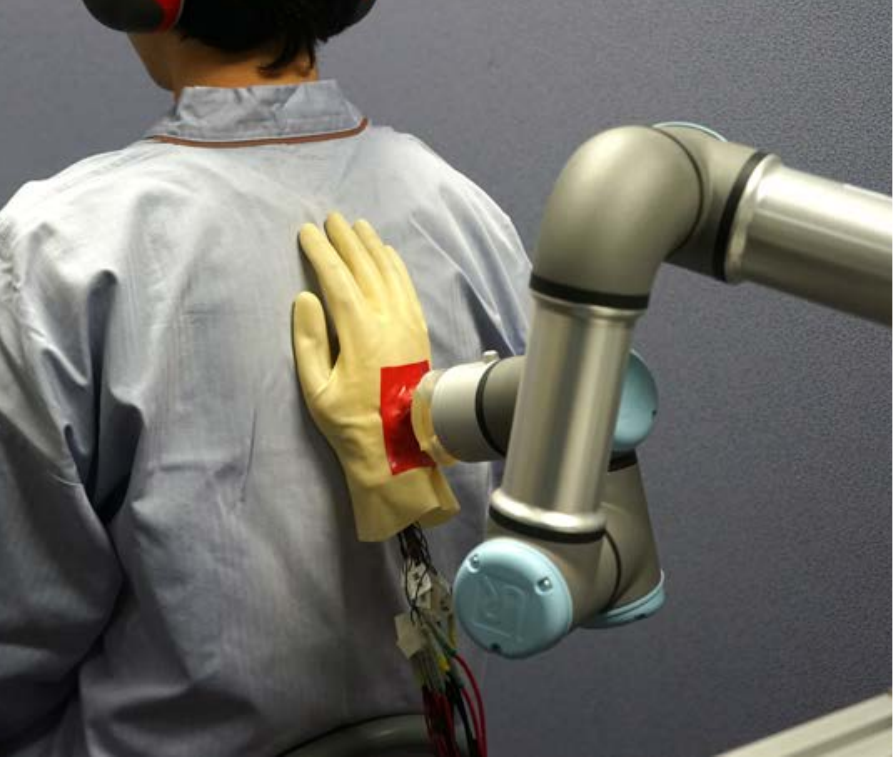}
         %\hspace{1.6cm} A. Robot stroke motion.
        \end{center}
      \end{minipage}
      % 2
      \begin{minipage}{0.45\hsize}
        \begin{center}
          \includegraphics[keepaspectratio=true, width=1.0\columnwidth]{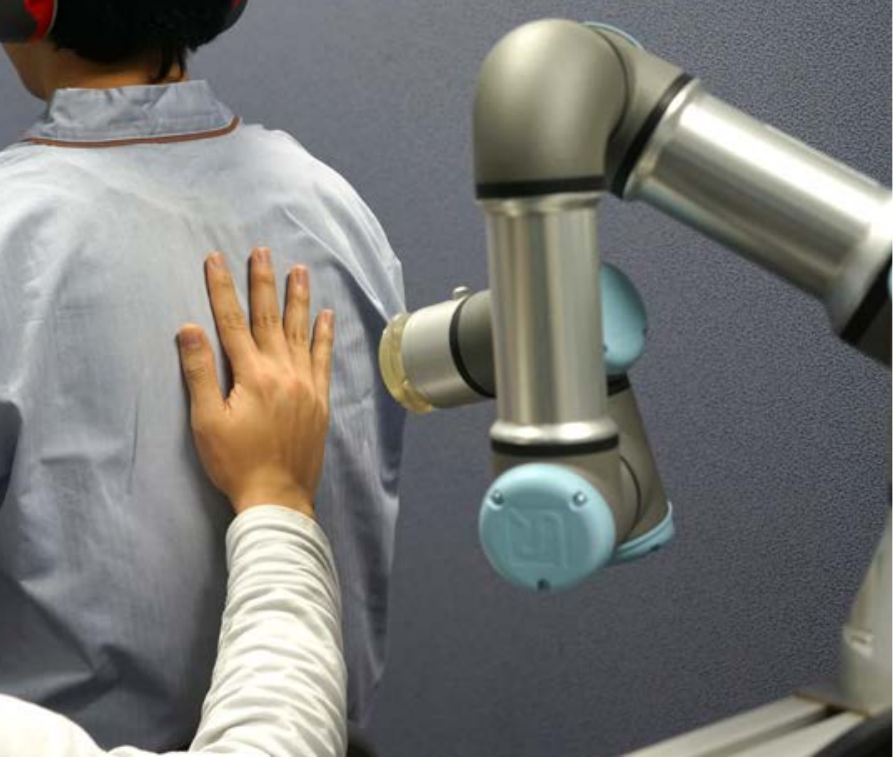}
          %\hspace{1.6cm} B. Human stroke motion.
        \end{center}
      \end{minipage}
    \end{tabular}
    \caption{Comparative experimental conditions for the robot stroke motion (left) and the human stroke motion (right).}
    \vspace{-2mm}
    \label{overview}
  \end{center}
\end{figure}
%%%%%%%%%%%%%%%%%%%%%%%%%%%%%%%%%%%%%%%%%%%%%%%%

% For solving the first issue, we develop a human-imitation hand to mimic the rigidity of a human hand.
% We believe that a human-like touch gives us a better feeling.
To solve the first issue, we developed a human-imitation hand to mimic the rigidity of a human hand—we believe that human-like touch feels better for users.
% Actually, we develop the robot system that gently strokes the back of a human to provide us a positive emotional effect, as shown in Fig.~\ref{overview}.
We developed a robot that gently stroked the back of a human to provide a positive emotional effect, as shown in Fig.~\ref{overview}. 
% For solving the second issue, we verify two speeds (3 cm/s and 30 cm/s~\cite{loken2009coding}) of stroking from subjects’ valence-arousal score.
% The score can measure a variety of emotions~({\ie} relaxation and excitement).
To solve the second issue, we verified two stroking speeds (3 cm/s and 30 cm/s~\cite{loken2009coding}) using the subjects’ valence-arousal score—the score being able to measure a variety of emotions (i.e., relaxation and excitement).

% In this paper, we further propose the method for evaluating the performance of affective touch by a robot.
In this paper, we propose a method for evaluating the performance of affective touch by a robot.
% Since our final goal is to realize robotic affective touch as a human does, we compare the gentle stroke by the proposed method with the stroke by a human.
Since our final goal was to realize robotic affective touch as close to humanly done as possible, we compared a gentle stroke using the proposed method with the stroke by a human.
% We carefully design the experimental protocols to compare them.
We carefully designed the experimental protocols to compare them.

% The contribution of this paper is twofold.
The contribution of this study is twofold.
% We propose and develop a human-imitation hand, which has human-like flexibility, softness, and warmness.
First, we proposed and developed a human-imitation hand having human-like flexibility, softness, and warmth. 
% Second, we compare the emotional effects against both a human hand and the imitation hand subjectively and physiologically.
Second, we compared the emotional effects on patients against both the human hand and the imitation hand, subjectively and physiologically. 
% Though affective touch by a robot is a little inferior to affective touch by a human, we find out three interest findings.
Though affective touch by a robot was a little inferior to human affective touch, we made three interesting findings.
% First, the subjects evaluate strokes at a similar tendency in the speed aspect with both human hand and human-imitation hand in both subjective and physiological evaluations.
First, the subjects evaluated strokes similarly with regard to the stroke speed of the human and human-imitation hand, in both the subjective and physiological evaluations. 
% Second, the subjects feel more pleasant and arousal with the faster-speed (30 cm/s than 3 cm/s) stroke.
Second, the subjects felt more pleasure and arousal at a faster stroke rate (30 cm/s rather than 3 cm/s). 
% Third, less fitting due to bending a back gets the bad effect and there is the room to enhance the system by improving the fitting mechanism of the proposed hand.
Third, poorer fitting of the human-imitation hand due to the bending of the back had a negative emotional effect on subjects, suggesting that there is room to enhance the system by improving the fitting mechanism of the proposed hand.

\section{Related Works}
% As improving the safety and hardware of robot technology, the distance between humans and robots in human-robot interaction (HRI) has become closer.
The distance between humans and robots in human–robot interaction (HRI) has become closer to improve the safety and hardware of robot technology.
% \textit{Paro}~\cite{wada2004paro} as a therapy robot interacts with a user through him/her touching and stroking its fluffy surface.
\textit{Paro}~\cite{wada2004paro}—a therapy robot—interacts with a user their touching and stroking its fluffy surface. 
% \textit{Telenoid}~\cite{ogawa2011telenoid} shows the effectiveness of the tactile sensing and stimuli in remote interaction through a user hugging.
\textit{Telenoid}~\cite{ogawa2011telenoid} shows the effectiveness of tactile sensing and stimuli in remote interaction through user hugging.
% In these research methods, a subject actively contacts a robot for giving a positive impression.
In these research methods, the subject actively makes contact with a robot to convey a positive impression.

% Since humans both touch to and are touched from the others in physical human-to-human interaction, {\em active touch} by robots~({\eg} a robot attempts to touch humans) is also important.
Since humans both touch and are touched by others in physical human-to-human interaction, {\em active touch} by robots ({\eg} a robot attempting to touch a human) is also important. 
% To realize to stroke the human body using a robot arm in physical human-robot interaction (pHRI), several studies have been attempted.
Several studies have examined the use of a robot arm to stroke the human body in physical human–robot interaction (pHRI).
% King~\textit{et~al.}~\cite{king2010} developed a robot system to wipe the upper arm of a patient who lies on a bed by using a laser range finder and a dual arm robot.
King~\textit{et~al.}~\cite{king2010} developed a robot system to wipe the upper arm of a patient lying on a bed by using a laser range finder and a dual-arm robot. 
% Chen~\textit{et~al.}~\cite{chen2011} investigated the emotional effect by stroking a subject using the above robot~\cite{king2010}.
Chen~\textit{et~al.}~\cite{chen2011} investigated the emotional effect of stroking a subject using the above robot~\cite{king2010}.
% They indicated that the mechanical factors of the robot, such as appearance, design of the robot’s end-effector, and behavior, might provide a negative effect on the affective stroke.
They indicated that the mechanical factors of the robot such as its appearance, the design of the robot’s end-effector, and its behavior might have a negative effect on the benefits of the affective stroke.
% They suggest that the design of the robot’s end-effector needs to human-likeness such as a rubbery material and a warming element, and the physical and psychological factors of a robot need to be considered.
They suggested that the design of the robot’s end-effector needs to be human-like—perhaps using a rubbery material—and include some kind of warming element.
In short, the physical and psychological factors of a robot need to be considered.

% In a research of the relationship between the stroking speed and emotion, L{\"o}ken~\textit{et~al.}~\cite{loken2009coding} found that a firing pattern of low-threshold unmyelinated mechanoreceptors (C-tactile) correlated with ratings of pleasantness, and C-tactile responded most vigorously to 1-10 cm/s brushing, which gives us the most pleasant feeling.
In a study on the relationship between the stroking speed and emotion, L{\"o}ken~\textit{et~al.}~\cite{loken2009coding} found that a firing pattern of low-threshold unmyelinated mechanoreceptors (C-tactile fibers) correlated with ratings of pleasantness—the C-tactile fibers responded most vigorously to brushing at a rate of 1–10 cm/s, which produces the most pleasant feelings in participants. 
% Whether this stroke speed was suitable for affective touch by a robot was out of scope.
Whether this stroke speed was suitable for affective touch by a robot was beyond the scope of the study.

% Affective touch is the skill required in \textit{Humanitude}~\cite{humanitude}, one style of care giving. 
Affective touch is the skill required in \textit{Humanitude}~\cite{humanitude}, a type of care giving. 
% Based on the knowledge of Humanitude, Honda~\textit{et~al.}~\cite{honda2019HAI} indicated that the combination of stroke and speech rate have a correlation to provide comfort.
Based on the knowledge of \textit{Humanitude}, Honda~\textit{et~al.}~\cite{honda2019HAI} indicated that the combination of stroke and speech rate had a correlation with the provision of comfort.
% To make more human-like comfort, they suggest that the motion of stroke needs to be a more human-like stroke.
To offer more human-like comfort, they suggested that the motion of the stroke needed to be more human-like. 

% For affective touch devices, several studies on robot hands have been conducted.
Subsequently, several studies on robot hands for affective touch devices have been conducted.
% Nakanishi~\textit{et~al.}~\cite{nakanishi2014} developed a robot hand with a holding mechanism and a similar warmness as the human body to realize a handshake. 
Nakanishi~\textit{et~al.}~\cite{nakanishi2014} developed a robotic hand with a holding mechanism and a similar warmness to the human body to accomplish a handshake.
% The proposed hand provided the feeling of human-likeness, but the hand was applied to handshaking only.
The proposed hand provided the feeling of human likeness, but the hand was applied to handshaking only.
% Cabibihan~\textit{et~al.}~\cite{cabibihan2015} developed a warm and soft artificial hand that gives the illusion of feeling like a human touch.
Cabibihan~\textit{et~al.}~\cite{cabibihan2015} developed a warm and soft artificial hand that gave the illusion of feeling like the touch of a human.
% The hand had silicon and a skeletal structure to realize a softness like a human hand and created an illusion that the touch is from a human hand.
The hand was made from silicon and had a skeletal structure to create a human-like softness and the illusion of being touched by a human hand.
% Ueno~\textit{et~al.}~\cite{ueno2019ROBIO} developed an artificial human-mimetic hand and forearm to reproduce human-to-human physical contact.
Ueno~\textit{et~al.}~\cite{ueno2019ROBIO} developed an artificial human mimetic hand and forearm to reproduce human-to-human physical contact.
% Also, Ueno~\textit{et~al.}~\cite{ueno2020ROMAN} shake hands with a highly realistic artificial hand and a manipulator.
In addition, Ueno~\textit{et~al.}~\cite{ueno2020ROMAN} shook hands with a highly realistic artificial hand and manipulator.
% The main concern of these methods is to design robot hands and the actual effect of emotion by affective touch by the hands was not evaluated yet.
The main challenge associated with these methods is that of the robot hands design.
Moreover, the actual emotive effect caused by the affective touch of these hands has not yet been evaluated.

%%%%%%%%%%%%%%%%%%%%%%%%%%%%%%%%%%%%%%%%%%%%%%%%
\begin{figure}[t]
\vspace{2mm}
  \centering
        \includegraphics[keepaspectratio=true, width=1.0\columnwidth]{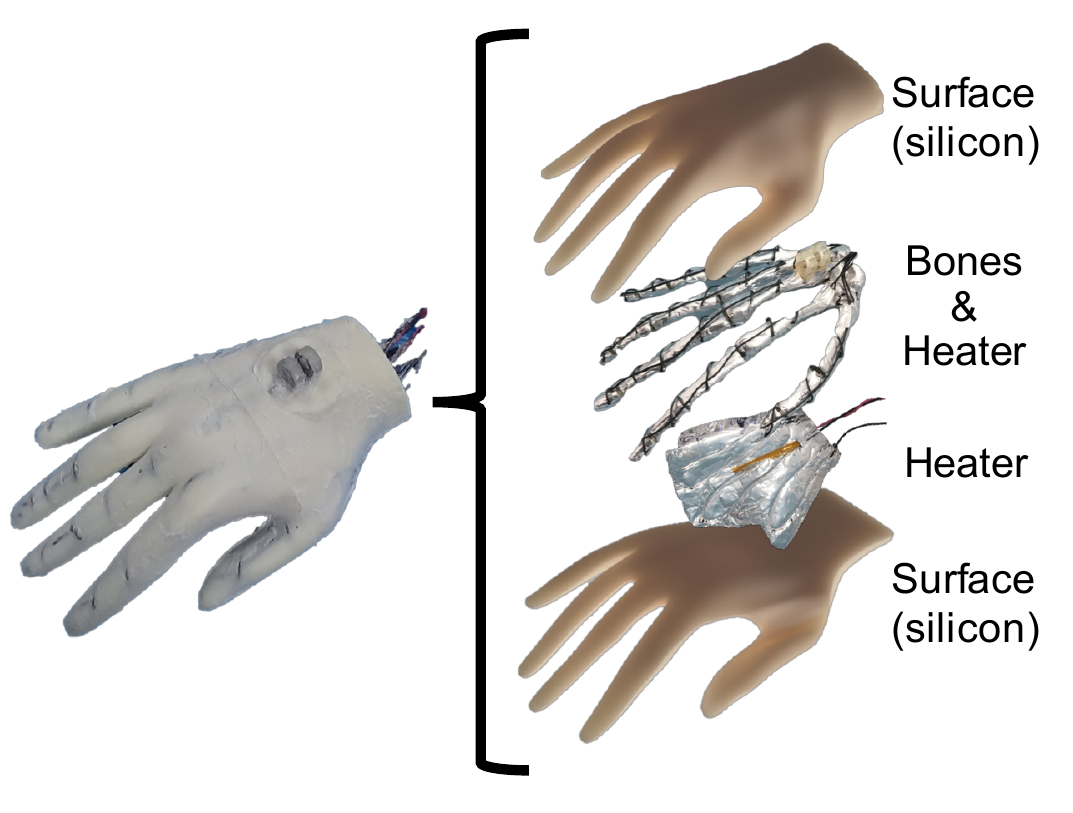}
        \caption{The structure of the proposed human-imitation hand.}
        \label{IHH}
\end{figure} 
%%%%%%%%%%%%%%%%%%%%%%%%%%%%%%%%%%%%%%%%%%%%%%%%

\section{Gentle Stroke by a Robot}
\subsection{Human-Imitation Hand}
% Toyoshima~\textit{et~al.}~\cite{toyoshima2018WELCARO} found that the following two elements are important to creating a robot hand for affective touch: 1) to fit the stroking surface; 2) to get warmed similar to a human hand.
Toyoshima~\textit{et~al.}~\cite{toyoshima2018WELCARO} found that the following two elements were important for creating a robotic hand for affective touch. 
The need for the robotic hand to:
\begin{enumerate}
\renewcommand{\labelenumi}{\arabic{enumi})}
 \item Fit the stroking surface;
 \item Feel as warm as a human hand.
\end{enumerate}

% As described in Section~\ref{intro}, following the two elements, we develop a human imitation hand, not a mechanical hand. The developed hand satisfies the following three aspects:
As described in Section~\ref{intro}, we developed a human-imitation hand, rather than a mechanical hand, based on the above two elements.
The developed hand satisfied the following three criteria:
\begin{itemize}
 \item Flexibility: to naturally conform to the surface being touched;
 \item softness and stiffness similar to those of a human hand;
 \item and a warmness similar to that of a human hand.
\end{itemize}

\begin{figure}[t]
\vspace{2mm}
  \centering
        \includegraphics[keepaspectratio=true, width=1.0\columnwidth]{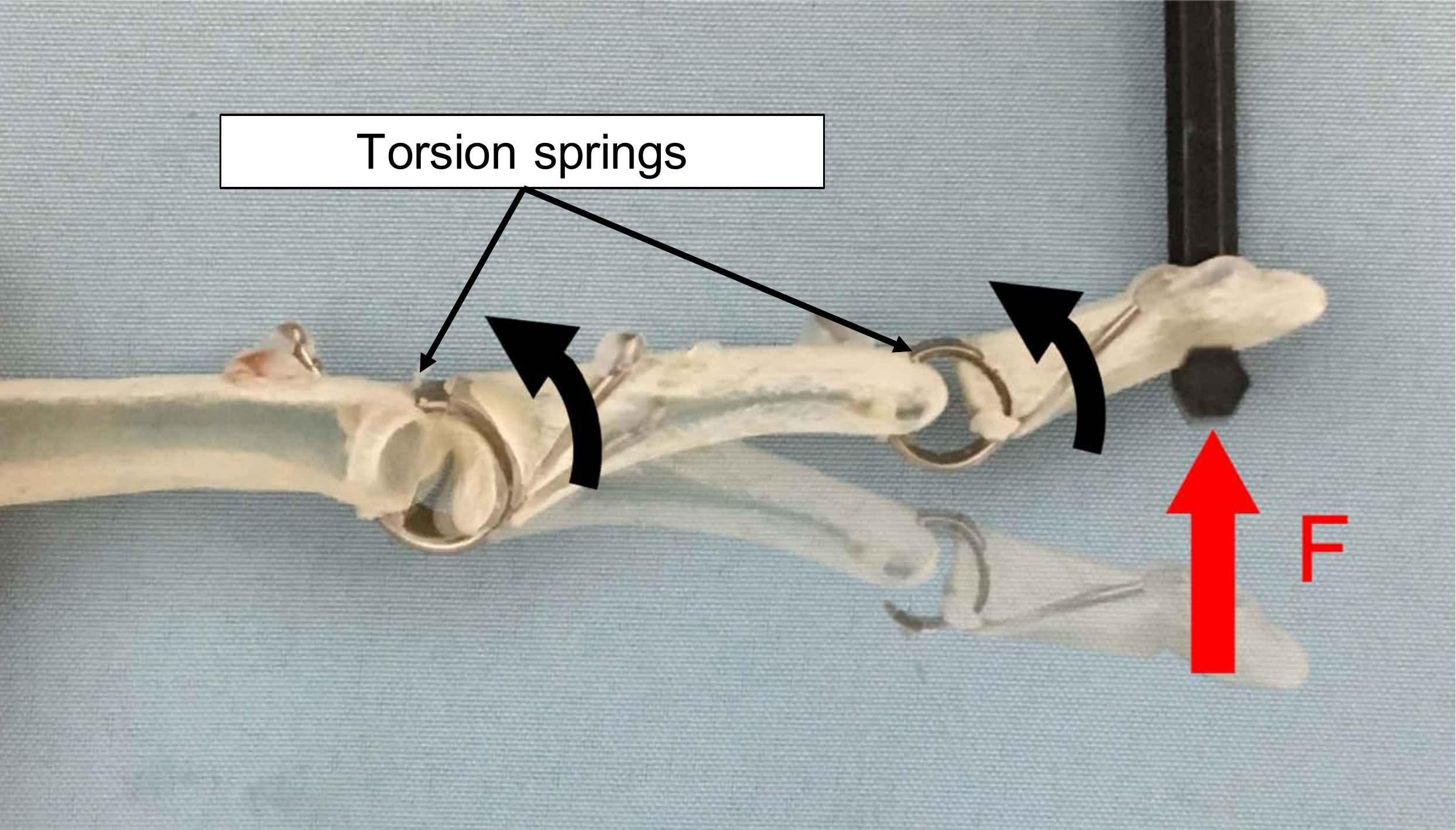}
        \caption{Transformation of the thumb with two passive joints.}
        \label{finger}
\end{figure}

% As shown in Fig.~\ref{IHH}, the proposed human-imitation hand consists of bones, joints, heaters (nichrome wires, film-type thermistor sensors, and aluminum foils), and material imitating body-tissue that provides softness while the inside bone provides the stiffness similar to a human hand.
As shown in Fig.~\ref{IHH}, the proposed human-imitation hand consisted of bones, joints, heaters (nichrome wires, film-type thermistor sensors, and aluminum foils), a body-tissue material coating that provided softness, and internal bones that provided a stiffness similar to that of a human hand.
% The bones are made with 3D printed material~(\textit{AR-M2}, KEYENCE CORPORATION\footnote{AR-M2, KEYENCE CORPORATION, https://www.keyence.co.jp/products/3d-printers/3d-printers/agilista-3100/models/ar-m2/}).
The bones were made using 3D printed material~(\textit{AR-M2}, KEYENCE CORPORATION\footnote{AR-M2, KEYENCE CORPORATION, https://www.keyence.co.jp/products/3d-printers/3d-printers/agilista-3100/models/ar-m2/}).
% The shape of bones is a human skeleton model from \textit{STLFinder}\footnote{STLFinder, https://www.stlfinder.com/}.
The shape of the bones was based on a human skeleton model from the \textit{STLFinder}\footnote{STLFinder, https://www.stlfinder.com/}.
% As shown in Fig.~\ref{finger}, between the bones, the torsion springs (0.3 Nmm/degree) are used as the joints.
As shown in Fig.~\ref{finger}, torsion springs (0.3 Nmm/degree) between the bones were used as joints. 
% The spring coefficient is decided from the finding of the related work~\cite{toyoshima2018WELCARO}.
The spring coefficient was determined from the findings of related work~\cite{toyoshima2018WELCARO}.
% We cover the bones and joints with aluminum foils, and wind nichrome wires around the bones, attach film-type thermistor sensors (temperature sensors) to uniformly warm the human-imitation hand.
We covered the bones and joints with aluminum foil, wound nichrome wires around the bones and attached film-type thermistor sensors (temperature sensors) to uniformly warm the human-imitation hand. 
% We use a type of soft silicone with a hardness of asker C7 (\textit{HITOHADA GEL}, EXSEAL CO., Ltd.\footnote{HITOHADA GEL, EXSEAL Co., Ltd, http://www.exseal.co.jp/english/\\creative/hitohada.html}) which has softness similar to human skin to cover the bones, joints, and heaters.
We used a type of soft silicone with a hardness of Asker C7~(\textit{HITOHADA GEL}, EXSEAL CO., Ltd.\footnote{HITOHADA GEL, EXSEAL Co., Ltd, http://www.exseal.co.jp/english/\\creative/hitohada.html}) which has a softness similar to that of human skin to cover the bones, joints, and heaters.
% Finally, we use a natural rubber glove to cover and protect the silicon.
Finally, we used a natural rubber glove to cover and protect the silicon.

% We control current of nichrome wires to keep the temperature of the human-imitation hand with temperature sensors.
We controlled the current in the nichrome wires to maintain the temperature of the human-imitation hand using temperature sensors.
% To reduce the burden of each heater, we use six nichrome wire - temperature sensor pairs and warm six parts of the hand independently.
To reduce the burden on each heater, we used six nichrome wire-temperature sensor pairs and warmed six parts of the hand independently.
% The six parts are the thumb, index, middle, ring, pinky, and palm.
The six regions were the thumb, index finger, middle finger, ring finger, little finger, and the palm.
% A target values of each part respectively are set to 49, 48, 46, 44, 44, and 42~\(^\circ\)C to keep the surface temperature of the human-imitation hand at about 35~\(^\circ\)C by a PID controller.
The target values of each part were set to 49, 48, 46, 44, 44, and 42~\(^\circ\)C to maintain the surface temperature of the human-imitation hand at approximately 35~\(^\circ\)C using a PID controller.

\begin{figure}[t]
  \begin{center}
    \begin{tabular}{c}
    
      % 2
      \begin{minipage}{0.5\hsize}
        \begin{center}
          \includegraphics[keepaspectratio=true, width=1.0\columnwidth]{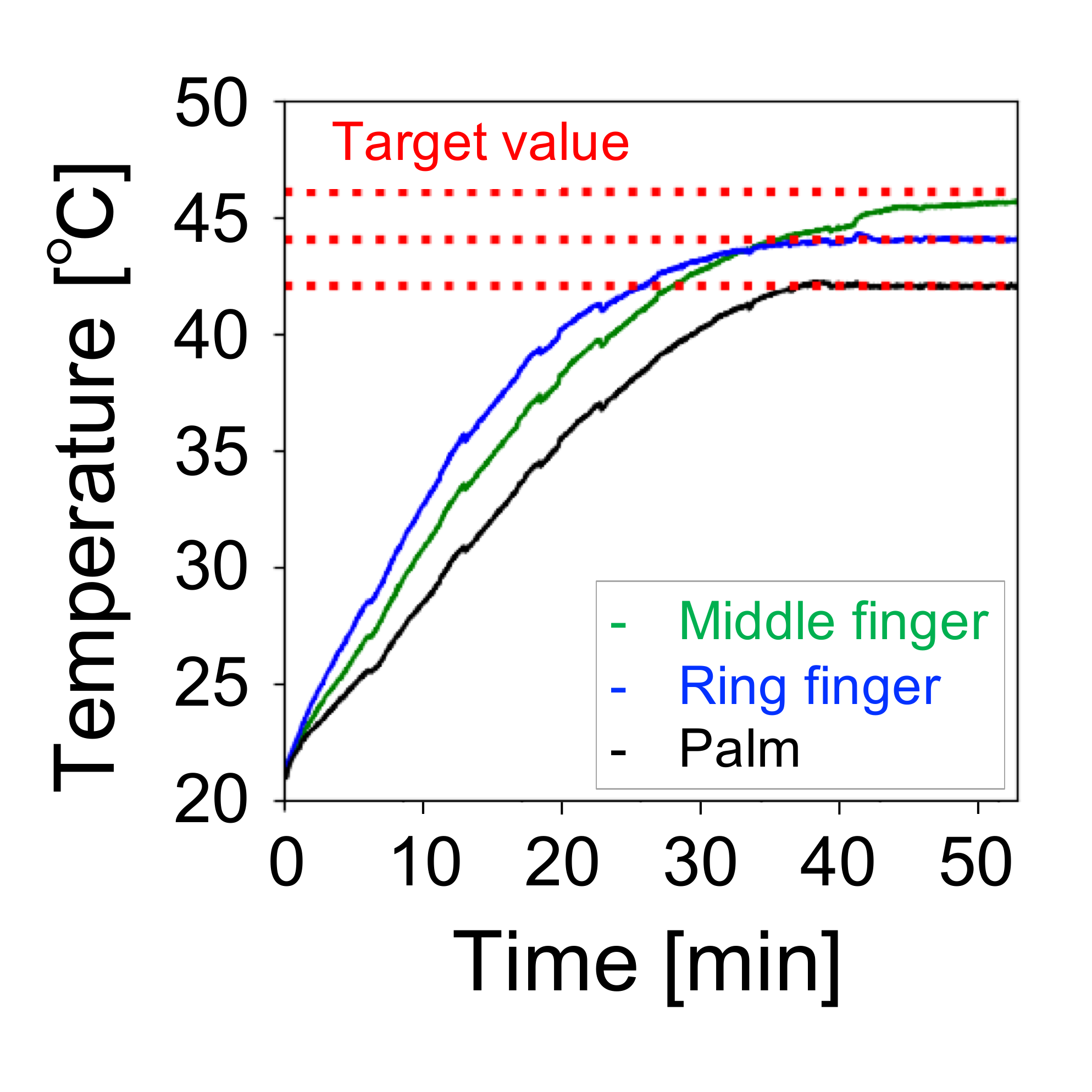}
          \hspace{1.6cm} A. The temperature graph.
        \end{center}
      \end{minipage}

      % 1
      \begin{minipage}{0.5\hsize}
        \begin{center}
          \includegraphics[keepaspectratio=true, width=1.0\columnwidth]{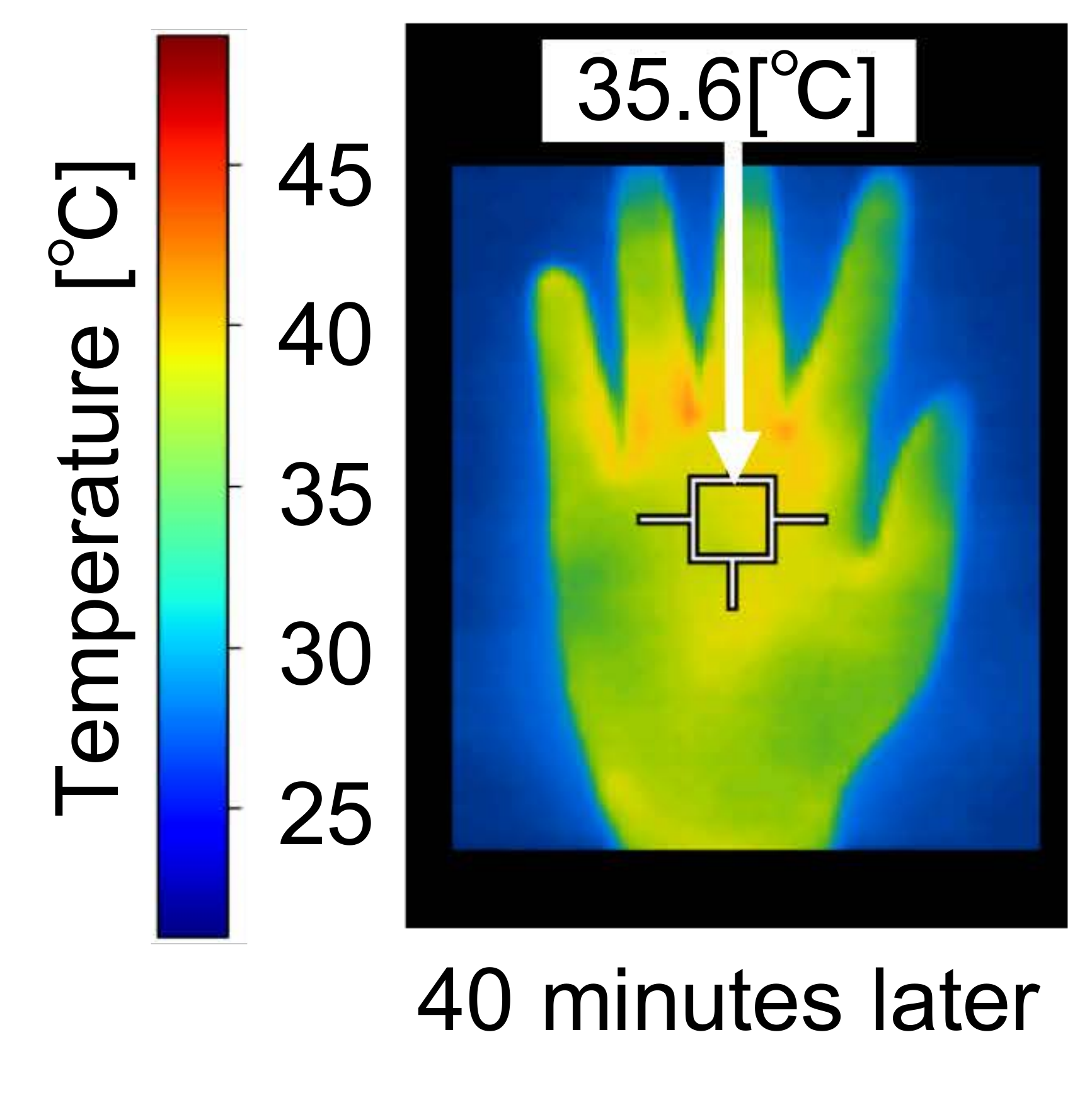}
         \hspace{1.6cm} B. The thermographic image.
        \end{center}
      \end{minipage}

    \end{tabular}
    \caption{The temperature of the imitational hand.}
    \label{handtemperature}
  \end{center}
\end{figure}

% We evaluate the performance of the temperature control of the hand.
We evaluated the performance of the temperature control of the hand.
% Fig.~\ref{handtemperature} (A) shows the values of temperature sensors during temperature control for 50 minutes. Each value of temperature sensor is achieved the target value at around 40 minutes.
Fig.~\ref{handtemperature} (A) shows the values of the temperature sensors during the temperature control process for 50 min.
% Fig.~\ref{handtemperature} (B) shows the thermographic image of the surface of the hand.
Fig.~\ref{handtemperature} (B) shows a thermographic image of the surface of the hand.
% It is confirmed that the surface temperature of the hand was about 35°C, although the locally surface temperature has variation.
It was confirmed that the surface temperature of the hand was approximately 35 °C, although the local surface temperature varied.
% Based on the above, we warm up 40 minutes before the start of the experiment to use the human imitation hand.
Based on the above, we warmed up the human-imitation hand 40 min before the start of the experiment.

\subsection{Generation of Stroke Motion}
\label{stroke}
% We attach the proposed hand to the robot arm to stroke the back of a human.
We attached the proposed hand to the robot arm to stroke the back of a human. 
% It is very common to stroke the back for positive emotional effects in {\em tactile care}~\cite{andersson2009tactile}.
It is very common to stroke the back for positive emotional effects in {\em tactile care}~\cite{andersson2009tactile}.
% In addition, the subject cannot directly see the robot in this condition and this reduces the psychological influence by the appearance of the robot.
Moreover, the subject cannot directly see the robot in this scenario, which reduces the psychological influence of the robot’s appearance.

% Based on the knowledge of tactile care~\cite{koizumi2017}, the robot attempts to realize the gentle stroke whose pressure and speed keeps constant.
Based on its knowledge of {\em tactile care}~\cite{koizumi2017}, the robot attempts to execute a gentle stroke whose pressure and speed remains constant. 
% First, to realize the motion with constant pressure using the robot arm, we adopt the impedance control.
First, to achieve motion at a constant pressure using the robot arm, we adopted an impedance control approach.
% The robot used in this paper has the impedance controller provided by the manufacturer and we use it as is.
The robot used in this study had the impedance controller provided by the manufacturer, and we used it \textit{as is}.
% We set the target force to 3N following the suggestion in~\cite{chen2011}.
We set the target force to 3 N following the recommendation in~\cite{chen2011}.

% Second, it is easy to realize the motion with the constant speed using the robot motion, if we decide the speed.
Second, it is easy to achieve motion at a constant speed using the robot motion if we choose the speed itself. 
% We choose two types of speed (3 and 30 cm/s) following the related work~\cite{pawling2017c}, which proves that 3 cm/s stroke motion to an arm is more pleasant than a same number or a constant duration of strokes at 30 cm/s.
Consequently, we chose two speeds (3 and 30 cm/s) based on related work~\cite{pawling2017c}, which proved that a 3 cm/s stroking motion to an arm was more pleasant than the same number of strokes or a constant duration of the strokes at 30 cm/s. 
% In this paper, we compare the speeds with the same duration.
In this study, we compared the speeds over the same duration.

% Note that the spatial resolution is slightly different between the forearm and the back because of the difference in the two-point discrimination thresholds of the sensors~\cite{weinstein1968intensive}.
Note that the spatial resolution was slightly different between the forearm and the back because of the difference in the two-point discrimination thresholds of the sensors~\cite{weinstein1968intensive}. 
% This difference indicate that C-tactile distribution is different in the forearm and back.
This difference indicates that the C-tactile distribution was different in the forearm and back.

\begin{figure*}[t]
\vspace{2mm}
  \centering
        \includegraphics[keepaspectratio=true, width=2.04\columnwidth]{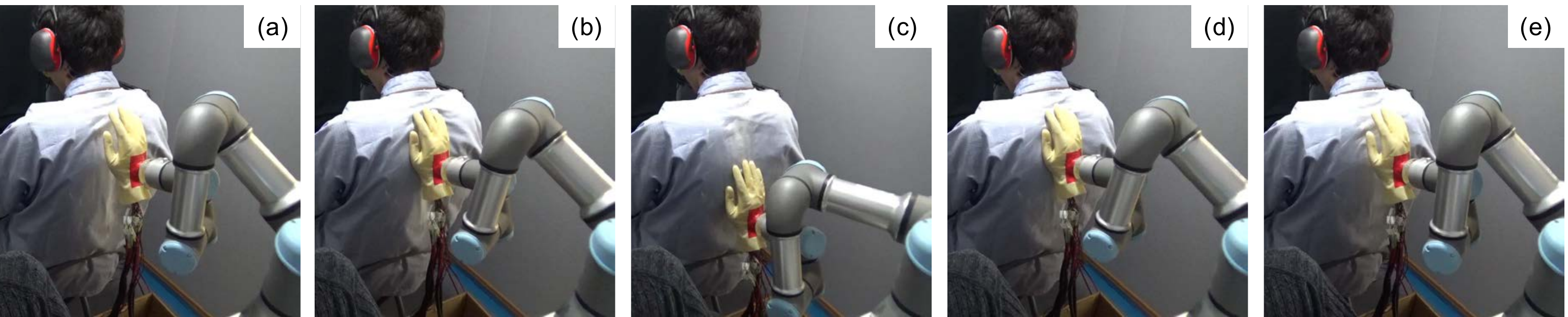}
        \caption{An example of the flow of the robot stroke motion. (a) Move towards the back. (b) Contact the back and keep pressure. (c)-(d) Stroke down and up the back. (e) Leave the back.}
        \label{robotstroke}
\end{figure*}

% Fig.~\ref{robotstroke} shows the flow to stroke the back using the robot arm.
Fig.~\ref{robotstroke} shows the flow to stroke the back using the robot arm. 
% First, the robot arm moves toward the back of a subject on 1 cm/s, and the robot arm keeps to press the middle of the subject's back.
First, the robot arm moved toward the back of a subject at 1 cm/s, and then pressed the middle of the subject’s back. 
% For removing a physiological effect caused by the initial contact, the robot keeps pressing for 10 seconds to wait for a subject to stay calm.
To minimize the physiological effect caused by the initial contact, the robot kept pressing for 10 s to help the subject remain calm. 
% Next, the robot arm moves down 15 cm and returns the touching position.
Next, the robot arm moved down 15 cm and returned to the touch position.
% The robot repeats the same motion for around 20 seconds.
The robot repeated the same motion for approximately 20 s. 
% After the robot arm finishes stroking, the robot arm takes to leave the back of the subject on 30 cm/s.
After the robot arm had finished stroking, it left the back of the subject at 30 cm/s.

%%%%%%%%%%%%%%%%%%%%%%%%%%%%%%%%%%%%%%%%%%%%%%%%
%%%%%%%%%%%%%%%%%%%%%%%%%%%%%%%%%%%%%%%%%%%%%%%%
%%%%%%%%%%%%%%%%%%%%%%%%%%%%%%%%%%%%%%%%%%%%%%%%
%%%%%%%%%%%%%%%%%%%%%%%%%%%%%%%%%%%%%%%%%%%%%%%%
%%%%%%%%%%%%%%%%%%%%%%%%%%%%%%%%%%%%%%%%%%%%%%%%
%%%%%%%%%%%%%%%%%%%%%%%%%%%%%%%%%%%%%%%%%%%%%%%%
\begin{figure*}[t]
\vspace{2mm}
  \centering
        \includegraphics[keepaspectratio=true, width=2.04\columnwidth]{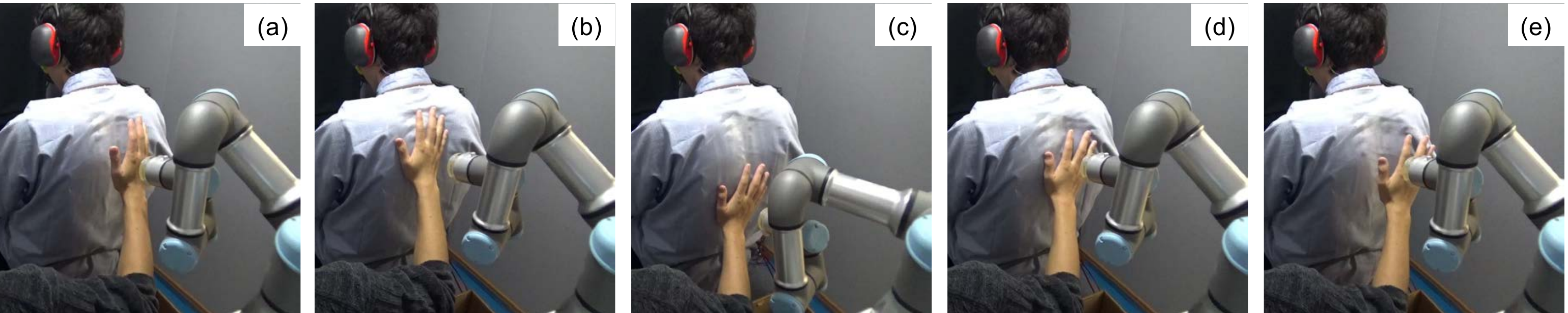}
        \caption{An example of the flow of the human stroke motion. (a) Move towards the back. (b) Contact the back and keep pressure. (c)-(d) Stroke the down and up the back. (e) Leave the back.}
        \label{humanstroke}
\end{figure*} 
%%%%%%%%%%%%%%%%%%%%%%%%%%%%%%%%%%%%%%%%%%%%%%%%

\section{Experimental Protocol}
\subsection{Experimental Purpose}
% To achieve the goal of this paper, ~{\ie} realization of human-like comfortable gentle stroke by a robot, the following two questions should be considered:
To achieve the aim of this paper, that is, the realization of a comfortable, gentle, and human-like stroke by a robot, the following two issues need to be considered:
% \begin{enumerate}
% \renewcommand{\labelenumi}{\arabic{enumi})}
% \item Slow stroking vs. rapid stroking. 
% We determined the suitable speed of the back stroke.
% Based on related work which found that slow stroking fired the C-tactile neurons (at 1–10 cm/s)~\cite{loken2009coding, pawling2017c} and tactile massage proved to be relaxating~\cite{tactilemassage}, we predicted that slow stroking would provide highly pleasurable, low-arousal feelings for the subjects, than rapid stroking.
% \item Human-imitation hand vs. human hand.
% We verified whether the human-imitation hand could provide a human-like touch by comparing it with a human hand.
% If a stroke by a robot could not be distinguished from a stroke by a human, both could evoke similar degrees of emotion.
% \end{enumerate}

1) Slow stroking vs. rapid stroking. We determined the suitable speed of the back stroke. Based on related work which found that slow stroking fired the C-tactile neurons (at 1–10 cm/s)~\cite{loken2009coding, pawling2017c} and tactile massage proved to be relaxating~\cite{tactilemassage}, we predicted that slow stroking would provide highly pleasurable, low-arousal feelings for the subjects, than rapid stroking.

2) Human-imitation hand vs. human hand. We verified whether the human-imitation hand could provide a human-like touch by comparing it with a human hand. If a stroke by a robot could not be distinguished from a stroke by a human, both could evoke similar degrees of emotion.

% -$Question$ $1$: Slow stroking vs. Rapid stroking. We determine the suitable speed of back stroke. 
% Based on related works that slow stroking provides firing of C-tactile by brushing (1-10 cm/s)~\cite{loken2009coding, pawling2017c} and relaxation in tactile massage~\cite{tactilemassage}, we predict that slow stroking provides high pleasant and low arousal for the subjects than rapid stroking.

% -$Question$ $2$: Human imitation hand vs. Human hand. We verify whether the human imitation hand provides a human-like touch by comparing it with a human hand.
% If stroke by a robot cannot be distinguished from stroke by a human, both can evoke the similar degree of emotion.

\subsection{Experimental conditions}
% To answer the questions, we consider two types of conditions.
To answer these questions, we considered two types of actions. 
% The first condition is the types of the agent's hand,~{\ie} the human-imitation hand and the human hand.
The first condition was the type of agent’s hand, that is, the human-imitation hand and the human hand. 
% The second condition is the types of speed.
The second condition was the speed.
% As we mentioned in Section~\ref{stroke}, we adopt the same conditions (3 and 30 cm/s) of the related work~\cite{pawling2017c}. 
As mentioned in Section~\ref{stroke}, we adopted the same conditions (3 and 30 cm/s) of the related work~\cite{pawling2017c}. 
% Table~\ref{condition} shows the conditions in the experiment.
Table~\ref{condition} shows these experimental conditions. 
% Therefore, we experiment with the four conditions to combine two types of the hand (the human-imitation hand and the human hand) and two types of the speed (3 and 30 cm/s) of movement.
Consequently, we experimented with the four conditions to combine two types of hands (the human-imitation hand and the human hand) and two speeds (3 and 30 cm/s).
%%%%%%%%%%%%%%%%%%%%%%%%%%%%%%%%%%%%%%%%%%%%%%%%
\begin{table}[t]
\vspace{2mm}
  \centering
    \caption{Four conditions in the experiment.}
    \begin{tabular}{|c|c|c|c|}\hline
    \multicolumn{2}{|c|}{} & \multicolumn{2}{c|}{Types of agent's hand} \\
    \cline{3-4}
    \multicolumn{2}{|c|}{} & Human imitation & Human\\
    \hline
    Speed & 3 & condition \#1 & condition \#3\\ 
    \cline{2-4}
    ($cm/s$) & 30 & condition \#2& condition \#4\\
    \hline
    \end{tabular}
        \label{condition}
\end{table}
%%%%%%%%%%%%%%%%%%%%%%%%%%%%%%%%%%%%%%%%%%%%%%%%
%%%%%%%%%%%%%%%%%%%%%%%%%%%%%%%%%%%%%%%%%%%%%%%%
\begin{figure*}[t]
  \begin{center}
    \begin{tabular}{c}
      % 1
      \begin{minipage}{0.5\hsize}
        \begin{center}
          \includegraphics[keepaspectratio=true, width=0.8\columnwidth]{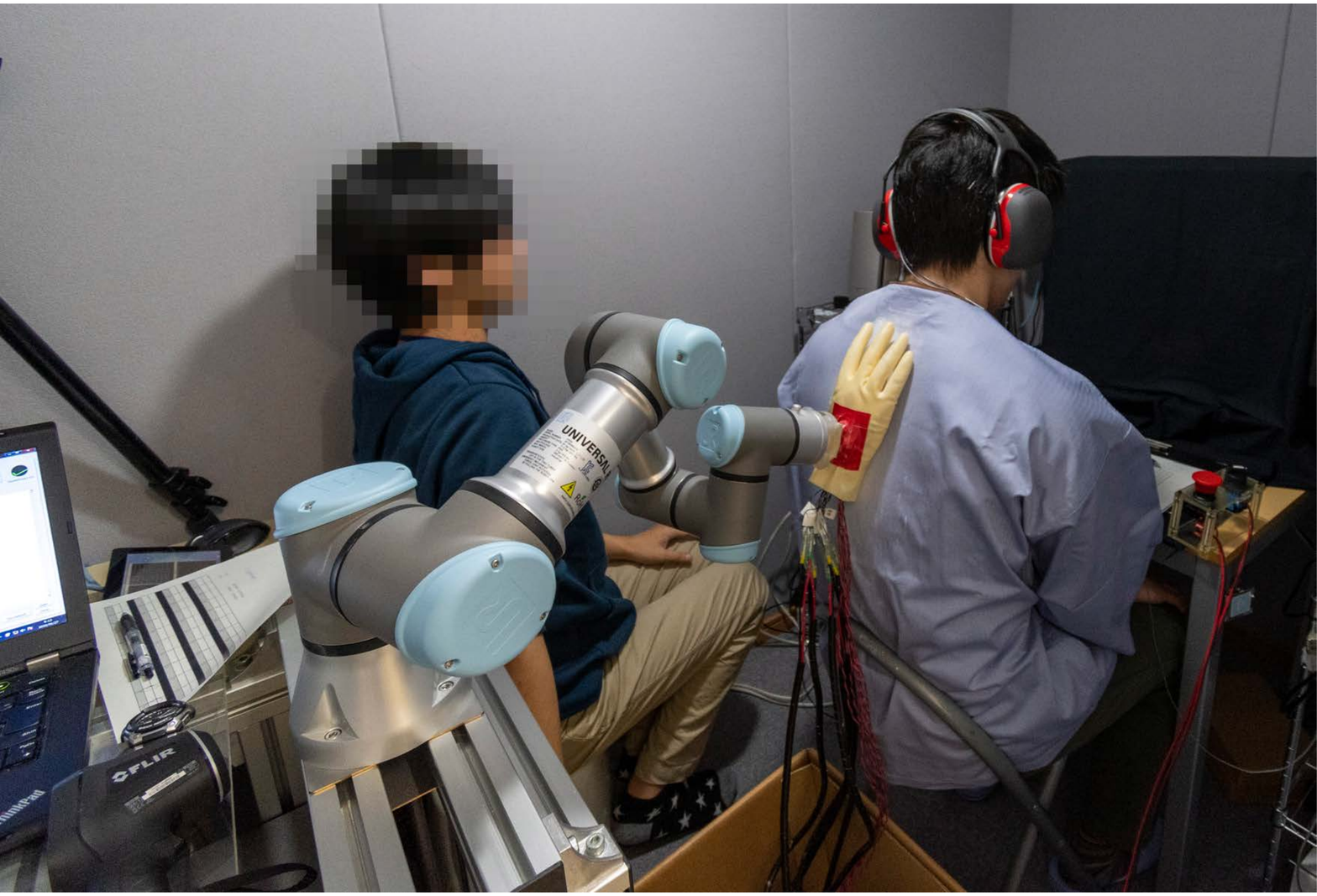}
         \hspace{1.6cm} A. The actual conditions of the experiment.
        \end{center}
      \end{minipage}

      % 2
      \begin{minipage}{0.5\hsize}
        \begin{center}
          \includegraphics[keepaspectratio=true, width=0.8\columnwidth]{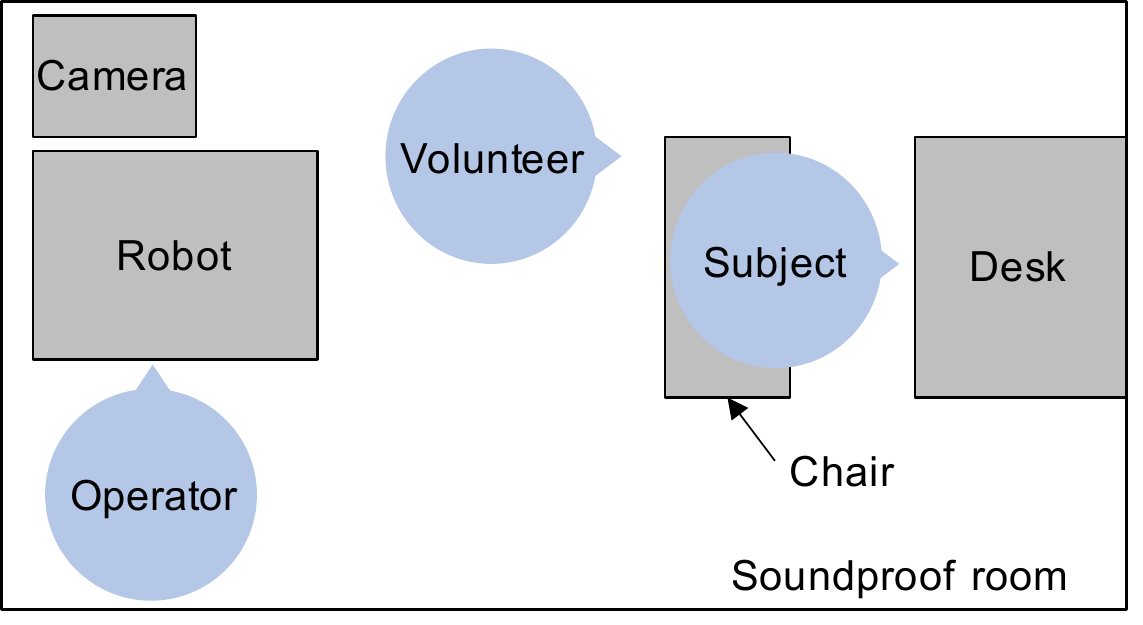}
          \hspace{1.6cm} B. Top view to show the orientational relation.
        \end{center}
      \end{minipage}
    \end{tabular}
    \caption{The room setting of the experiment.}
    \label{experimentroom}
  \end{center}
\end{figure*}
%%%%%%%%%%%%%%%%%%%%%%%%%%%%%%%%%%%%%%%%%%%%%%%%
%%%%%%%%%%%%%%%%%%%%%%%%%%%%%%%%%%%%%%%%%%%%%%%%
\begin{figure*}[t]
%\vspace{2mm}
  \centering
        \includegraphics[keepaspectratio=true, width=0.95\textwidth]{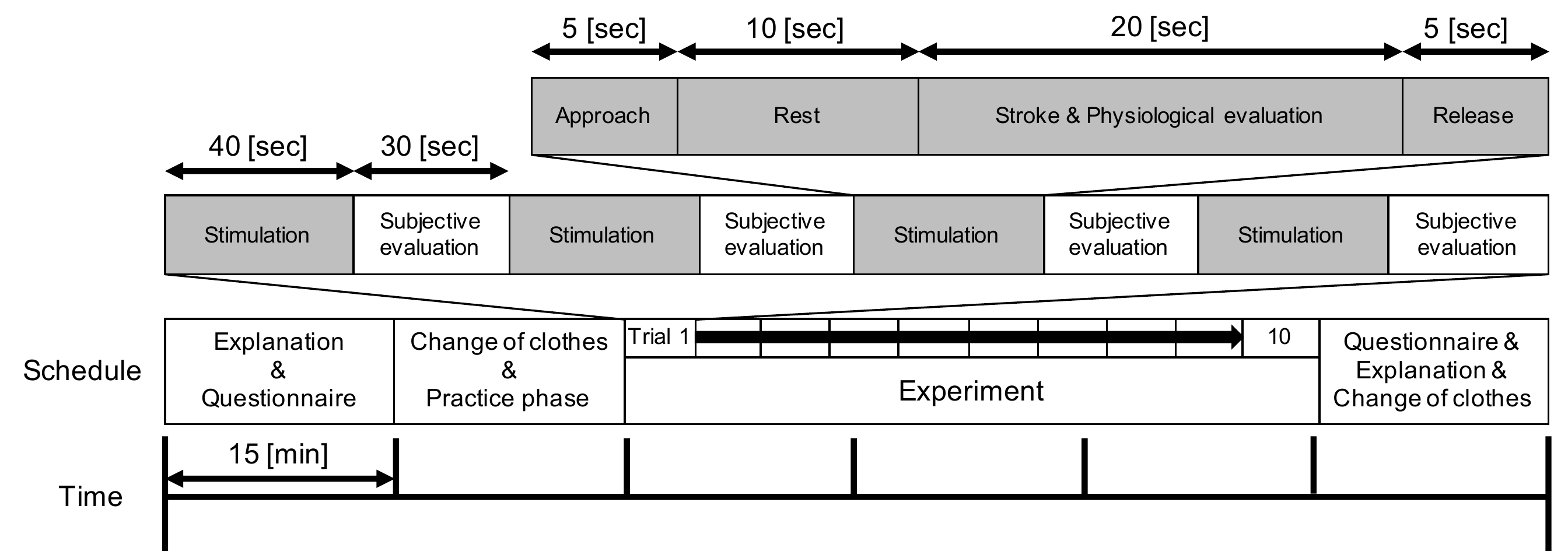}
        %\caption{The robot motion and the human motion.}
        \caption{The experimental flow.}
        \label{experiment_schedule}
\end{figure*} 
%%%%%%%%%%%%%%%%%%%%%%%%%%%%%%%%%%%%%%%%%%%%%%%%

\subsection{Experimental Control}
% In this experiment, we evaluate the emotional effect of the stroke by the proposed robot system compared to the stroke by a human.
In this experiment, we compared the emotional effect of the stroke by the proposed robot system with that of the stroke by a human.
% We recruited one volunteer (an adult male) for providing the human stroke and touch.
We recruited one volunteer (an adult male) to provide the human stroke and touch. 
% We demonstrate robotic stroke or human stroke randomly to a subject, and the subject evaluates it.
We demonstrated the robotic stroke or human stroke randomly to a subject, and the subject evaluated it.
% In order to encourage a subject to evaluate stroke touch feeling only, we remove other factors affecting the evaluation.
To encourage a subject to evaluate the feeling of the stroke touch only, we removed other factors affecting the evaluation.
% The worst scenario is that a subject notices the existence of another agent~({\eg} human) and evaluates it following the classification.
The worst-case scenario was if a subject should notice the existence of another agent~({\eg} human).
% We consider the following three factors that may provide the information to distinguish a robot and a human. 
We considered the following three factors that could provide information to distinguish between a robot and a human.

% First, we consider the sound caused by robot execution. 
First, we considered the sound caused by the operation of the robot.
% To handle this issue, we execute the robot fake motion even in the case of human stroke motion.
To handle this problem, we operated the robot (in simulated motion way) during the human stroke.
% The robot does not touch a subject but performs the same stroke motion as shown in Fig.~\ref{humanstroke}.
The robot did not touch the subject but performed the same stroke motion as shown in Fig.~\ref{humanstroke}. 
% Additionally, we ask a subject to wear an earmuff to reduce the sound from the outside.
Additionally, we asked the subject to wear earmuffs to reduce any outside sounds.

% Second, we consider the deviation of human stroke motions in each demonstration.
Second, we considered the deviation of the human stroke motions in each demonstration.
% As mentioned above, keeping speed and pressure is important.
As mentioned above, maintaining the speed and pressure is important. 
% To keep the speed, the robot guides the speed by showing the fake motion.
To maintain the correct speed, the robot guides the speed of the human hand by simulating the stroke motion (as mentioned above). 
% To keep the pressure, the volunteer trains to control the pressure by stroking a force plate and checking the measured force values in advance.
To maintain the pressure, the volunteer trained to control the pressure by stroking a force plate and checking the measured force values in advance.

% Third, we consider the possibility that the subjects imagine the agent other than the robot.
Third, we considered the possibility that the subjects imagine an agent other than the robot.
% To handle this issue, we tell the subjects that the robot provides all the strokes to the back.
To handle this problem, we told the subjects that the robot provided all back strokes. 
% Also, we tell the subjects that the volunteer keeps his eyes on the experiment to stop the robot in emergency. 
Moreover, we told the subjects that the volunteer would keep his eyes on the experiment to stop the robot in the event of an emergency. 
% Thus, we do not explicitly tell the subjects that the volunteer strokes the back.
Thus, we did not explicitly tell the subjects that the volunteer would stroke their back.
% Moreover, the subjects cannot see the agent in the experiment, they believe that the agent is only the robot.
Moreover, the subjects could not see the agent in the experiment; they believed that the only agent was the robot.

\subsection{Experimental Setup}
% The experiment is approved by both the ethics committee of the authors' affiliations.
The experiment was approved by both the ethics committee of the authors’ affiliations.
% Twelve subjects; adult males (20-25 years, Mean = 22.58 years, SD = 1.24) are recruited and obtained informed consent from all subjects after the explanation of experimental procedures.
Twelve subjects—adult males (20–25 years, mean = 22.58 years, SD = 1.24)—were recruited and informed consent was obtained from them all after the explanation of the experimental procedures.

% Fig.~\ref{experimentroom} shows the setup of the experimental room.
Fig.~\ref{experimentroom} shows the setup of the experimental room. 
% We place the collaborative robot arm~(\textit{UR3e}, Universal Robots\footnote{UR3e, Universal Robots, https://www.universal-robots.com/e-series/}), a chair on which the subject sits, a desk, and a camera in a soundproof room.
We placed the collaborative robot arm~(\textit{UR3e}, Universal Robots\footnote{UR3e, Universal Robots, https://www.universal-robots.com/e-series/}), a chair on which the subject sat, a desk, and a camera in a soundproof room.

% Fig.~\ref{experiment_schedule} shows the experiment flow.
Fig.~\ref{experiment_schedule} shows the experimental flow.
% First, before starting the experiment, the subject answers the individual information with a questionnaire. 
First, before starting the experiment, the subject answered an individual questionnaire.
% Next, to relieve the subject's anxiety caused by no experience of robot strokes, the subject takes a practice of the experience that the back of the subject is stroked by the robot at 3 cm/s.
Next, to relieve the subject’s anxiety caused by the lack of experience with robot strokes, the subject had a practice run of the experience—the robot stroked their back at 3 cm/s.

% After the practice, we repeat the following four processes 40 times:
After the practice run, we repeated the following 4 processes 40 times:
\begin{enumerate}
\renewcommand{\labelenumi}{\arabic{enumi})}
\item We randomly chose one of the four conditions and ran it on the subject; 
\item The subject rested for 10 s while being touched on the back to remove any physiological signals caused by the approach of the robot;
\item For the physiological evaluation, we measured the physiological signals of the subject for approximately 20 s of the stroke motion; and
\item The subject subjectively evaluated the stroke for 30 s.
\end{enumerate}
Finally, after the experiment, the subject answered a free-description type questionnaire to subjectively evaluate the experiment.

% 1) we randomly choose one of four conditions and run it to the subject; 
% 2) to remove the physiological signals of the subject caused by the approach of the robot, the subject rests for 10 seconds while being touched on the back;
% 3) we measure the physiological signals of the subject for the physiological evaluation during around 20 seconds of the stroke motion;
% 4) the subject subjectively evaluates the stroke for 30 seconds.
% Finally, after the experiment, the subject answers with a free-description type questionnaire to subjectively evaluate the experiment.

\subsection{Evaluation Indices}
% We use \textit{Affect Grid}~\cite{russell1989affect} for the subjective evaluation.
We used \textit{Affect Grid}~\cite{russell1989affect} for the subjective evaluation. 
% Parallelly, we adopt physiological evaluations related to subjective emotional responses~\cite{mayo2018putting,ree2019touch}, which are facial electromyogram (EMG) and a skin conductance level (SCL).
We also adopt physiological evaluations related to subjective emotional responses~\cite{mayo2018putting,ree2019touch}, such as facial electromyogram (EMG) and skin conductance levels (SCLs). 
% To support the evaluations, we survey the gentle stroke by the robot using a free-description type questionnaire.
To support the evaluations, we surveyed the gentle stroke by the robot using a free-description type questionnaire.

\subsubsection{\textit{Affect Grid}}
% \textit{Affect Grid} subjectively evaluates the emotional value.
\textit{Affect Grid} subjectively assesses emotional value.
% The subject plots two factors,~{\ie} valence and arousal, by placing a point on a 2D grid.
The subject plots two factors, that is, valence and arousal, by placing a point on a 2D grid.
% In the horizontal axis, from left to right, the unpleasant - pleasant degree is scored on nine scales as valence.
In the horizontal axis, from left to right, the unpleasant–pleasant degree is scored on nine scales as valence. 
% In the vertical axis, from top to bottom, the arousal - sleepiness degree is scored on nine scales as arousal.
In the vertical axis, from top to bottom, the arousal–sleepiness degree is scored on nine scales as arousal. 
% The areas of the upper right, the bottom right, the upper left, and the bottom left correspond to the states of excitement, relaxation, stress, and depression, respectively.
The areas of the upper right, bottom right, upper left, and bottom left correspond to the states of excitement, relaxation, stress, and depression, respectively. 
% We ask the subject to score valence-arousal by considering the deviation from the neutral state.
We asked each subject to score valence-arousal by considering the deviation from the neutral state.

\subsubsection{Facial EMG}
% The facial EMG are measured from a corrugator supercilii muscle (Cor) and a zygomatic major muscle (Zyg) for emotional valence.
The facial EMG is measured from the corrugator supercilii muscle (Cor) and a zygomatic major muscle (Zyg) for emotional valence.
% Basically, EMG activation of Cor and Zyg correspond to the unpleasant degree and the pleasant degree.
Essentially, EMG activation of the Cor and Zyg corresponds to an unpleasant and pleasant degree. 
% In our experiment, following the guidelines of Fridlund~\textit{et~al.}~\cite{fridlund1986guidelines} , Ag/AgCl electrode pads are attached at directly above the subject's left Cor and left Zyg.
Following the guidelines of Fridlund~\textit{et~al.}~\cite{fridlund1986guidelines}, Ag/AgCl electrode pads were attached directly above the subject’s left Cor and left Zyg.
% As ground electrode, the electrode pad is attached below the hairline in the center of the subject's forehead.
The electrode pad was attached below the hairline in the center of the subject’s forehead (as the ground electrode).

% We use the PowerLab 16/35 data acquisition system and LabChart Pro v8.0 software\footnote{ADInstruments, Dunedin, New Zealand} to collect data.
We used the PowerLab 16/35 data acquisition system and LabChart Pro v8.0 software\footnote{ADInstruments, Dunedin, New Zealand} to collect data.
% The data are amplified and sampled at 1000 Hz using an EMG-025 amplifier\footnote{Harada Electronic Industry, Sapporo, Japan}.
The data were amplified and sampled at 1000 Hz using an EMG-025 amplifier\footnote{Harada Electronic Industry, Sapporo, Japan}.
% As preprocessing, the data are filtered online (band-pass: 20–400 Hz).
For preprocessing, the data were filtered online (band-pass: 20–400 Hz). 

% Based on related research~\cite{sato2020facial}, the EMG data are analyzed by four steps.
Based on related research~\cite{sato2020facial}, the EMG data were analyzed in four steps.
% First, all EMG data are processed by full-wave rectification.
First, all EMG data were processed by full-wave rectification.
% Second, a baseline of EMG data in each trial is calculated as an average of EMG data from one second right before stroking.
Second, the baseline EMG data in each trial were calculated as an average of EMG data from one second just before stroking.
Third, the data during stroking is divided every one second and is calculated as the average of each section.
% Finally, the difference between each average and the baseline is calculated, and the results are summed to obtain an analysis data $x_i$.
Finally, the difference between each average and the baseline was calculated, and the results were summed to obtain the analysis data $x_i$. 

% For considering individual differences, the analysis data $x_i$ of all conditions in individuals are converted a standard score $z_i$ using eqaution (1), where $\overline{x}$ and $\sigma$ are calculated average and standard deviation in all conditions.
For consideration of individual differences, the analysis data $x_i$ of all conditions of all individuals were converted into a standard score $z_i$ using equation (1), where $\overline{x}$ and $\sigma$ are the calculated average and standard deviation in all conditions, respectively.
\begin{equation}
  z_i = \frac{x_i - \overline{x}}{\sigma}.
\end{equation}
The results are averages of all the standard scores $z_i$ of each condition.

\subsubsection{SCL}
% The SCL is measured from eccrine sweat gland activity.
The SCL was measured from eccrine sweat gland activity.
% Basically, active SCL and inactive SCL correspond to the arousal degree and the sleepiness degree.
Basically, active and inactive SCLs correspond to the degree of arousal and sleepiness.
% In our experiment, Ag/AgCl electrode pads are attached at the palmar surface of the middle phalanxes of the index and the middle fingers.
In our experiment, Ag/AgCl electrode pads were attached to the palmar surface of the middle phalanxes of the index and middle fingers.

% By applying a constant voltage of 0.5 V between the fingers, SCL is measured by Model 2701 BioDerm Skin Conductance Meter (UFI, Morro Bay, CA, USA).
By applying a constant voltage of 0.5 V between the fingers, the SCL was measured using a Model 2701 BioDerm Skin Conductance Meter (UFI, Morro Bay, CA, USA). 
% The measured data are sampled by the same equipment that measured EMG without the band-pass filter.
The measured data were sampled using the same equipment that measured the EMG without the band-pass filter.
% The SCL data are analyzed by the same method as EMG without full-wave rectification.
The SCL data were analyzed using the same method as the EMG without full-wave rectification.

%%%%%%%%%%%%%%%%%%%%%%%%%%%%%%%%%%%%%%%%%%%%%%%%
\begin{figure*}[t]
  \begin{center}
    \begin{tabular}{c}

      % 1
      \begin{minipage}{0.3\hsize}
        \begin{center}
          \includegraphics[keepaspectratio=true, width=1.0\columnwidth]{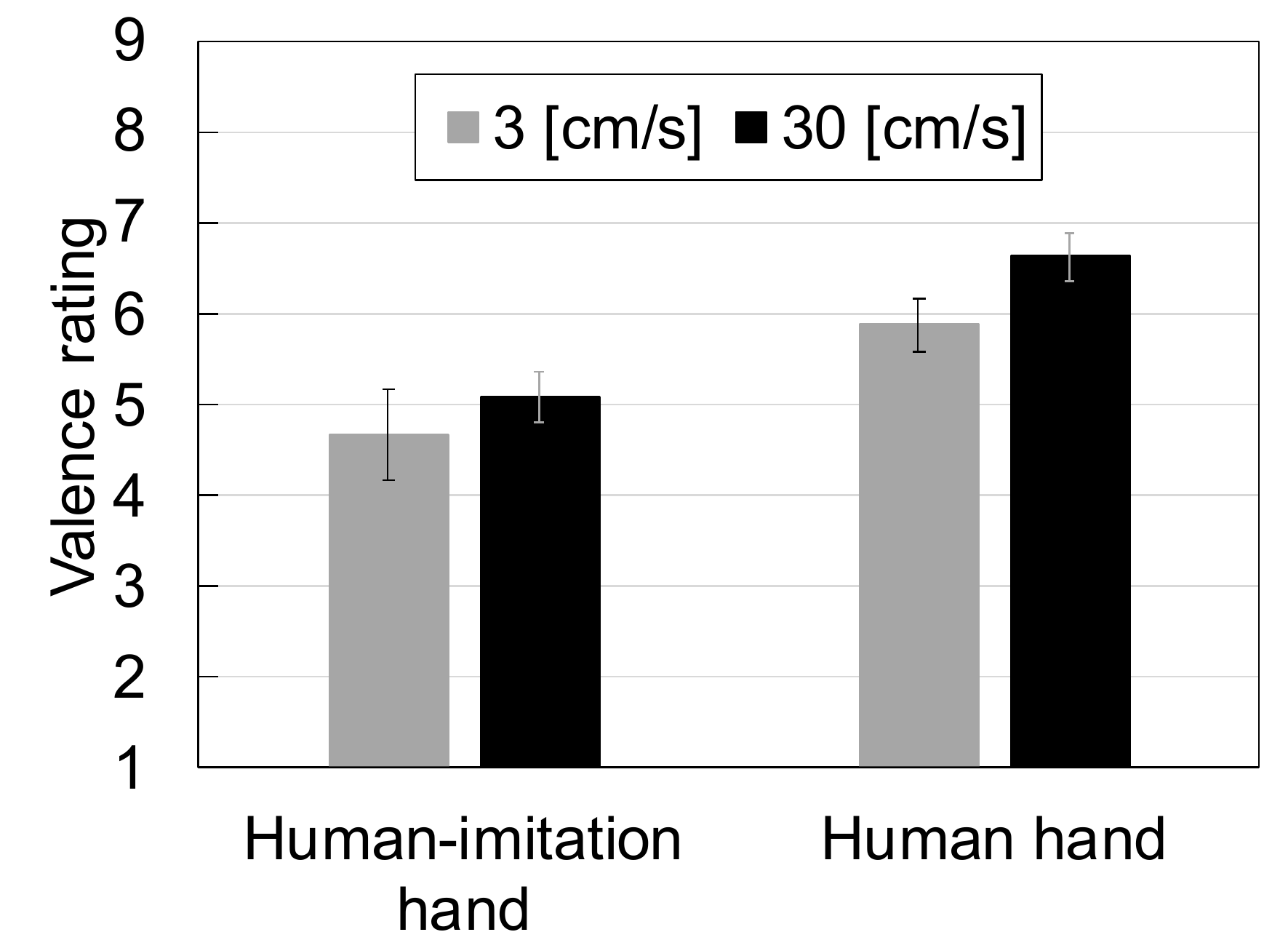}
         \hspace{1.6cm} A. Valence.
        \end{center}
      \end{minipage}

      % 2
      \begin{minipage}{0.3\hsize}
        \begin{center}
          \includegraphics[keepaspectratio=true, width=1.0\columnwidth]{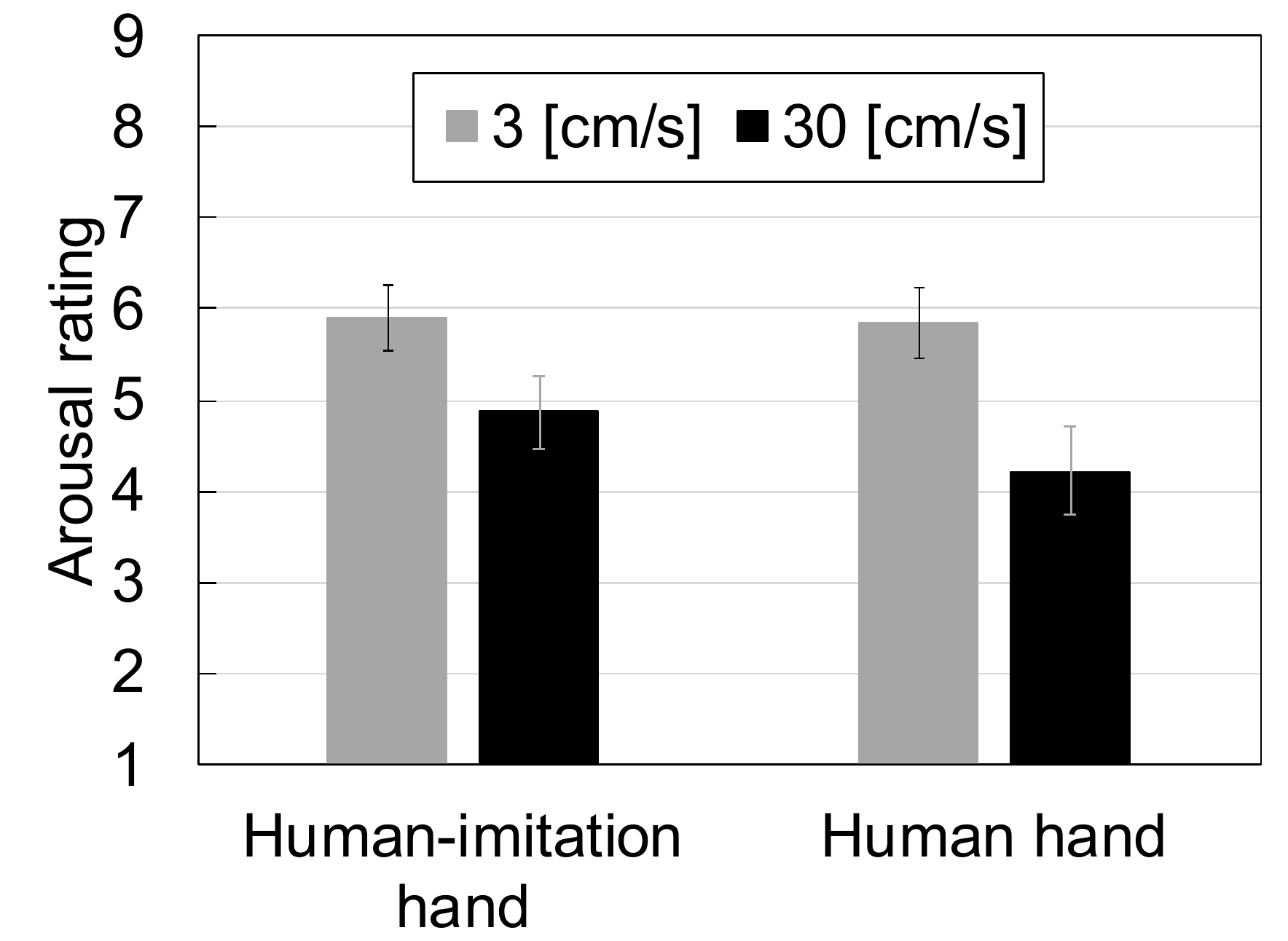}
          \hspace{1.6cm} B. Arousal.
        \end{center}
      \end{minipage}
    \end{tabular}
    \caption{The results of subjective evaluation.}
    \label{subjective}
  \end{center}
\end{figure*}
\begin{figure*}[t]
  \begin{center}
    \begin{tabular}{c}

      % 2
      \begin{minipage}{0.3\hsize}
        \begin{center}
          \includegraphics[keepaspectratio=true, width=1.0\columnwidth]{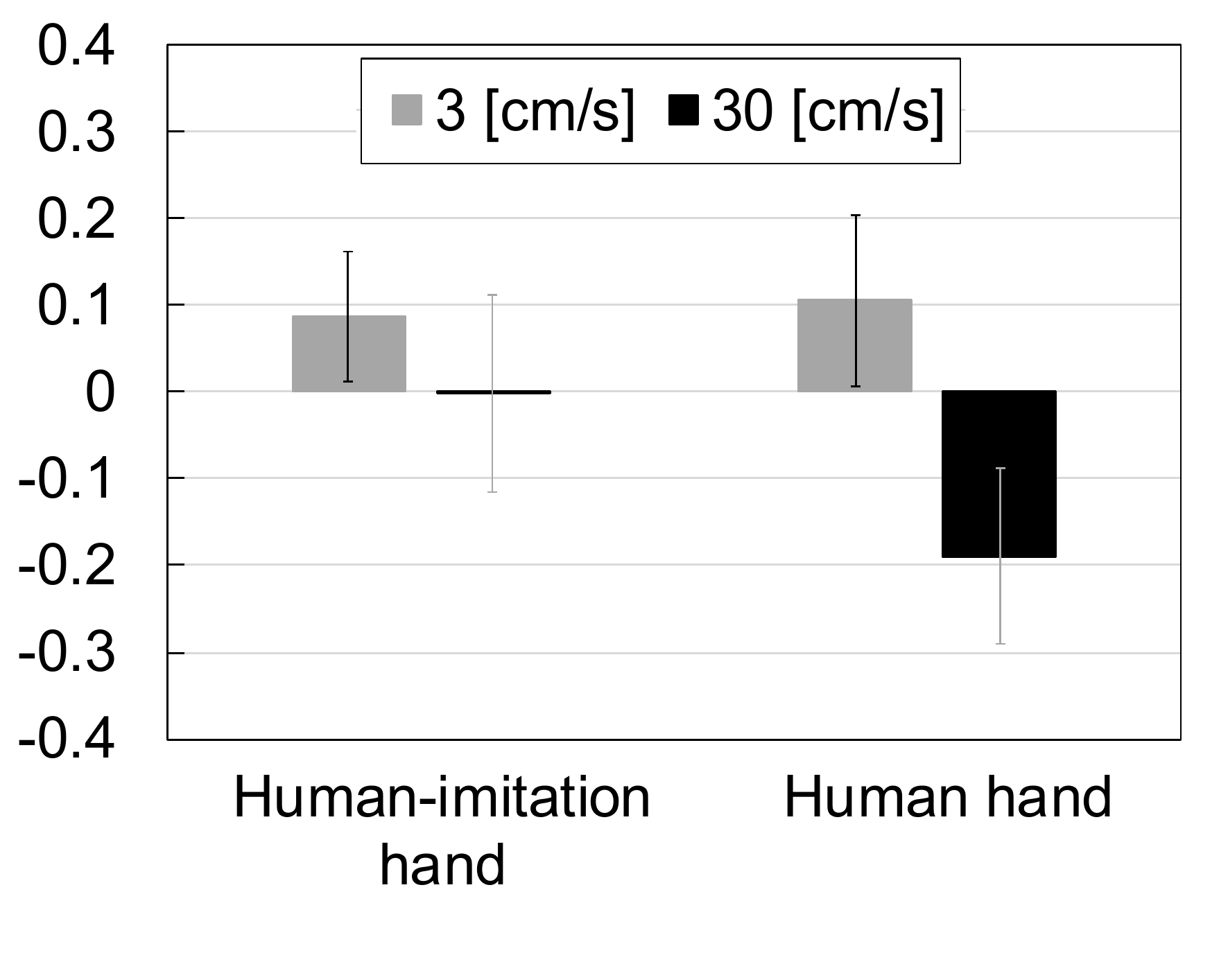}
          \hspace{1.6cm} A. Cor
        \end{center}
      \end{minipage}
      
      % 3
      \begin{minipage}{0.3\hsize}
        \begin{center}
          \includegraphics[keepaspectratio=true, width=1.0\columnwidth]{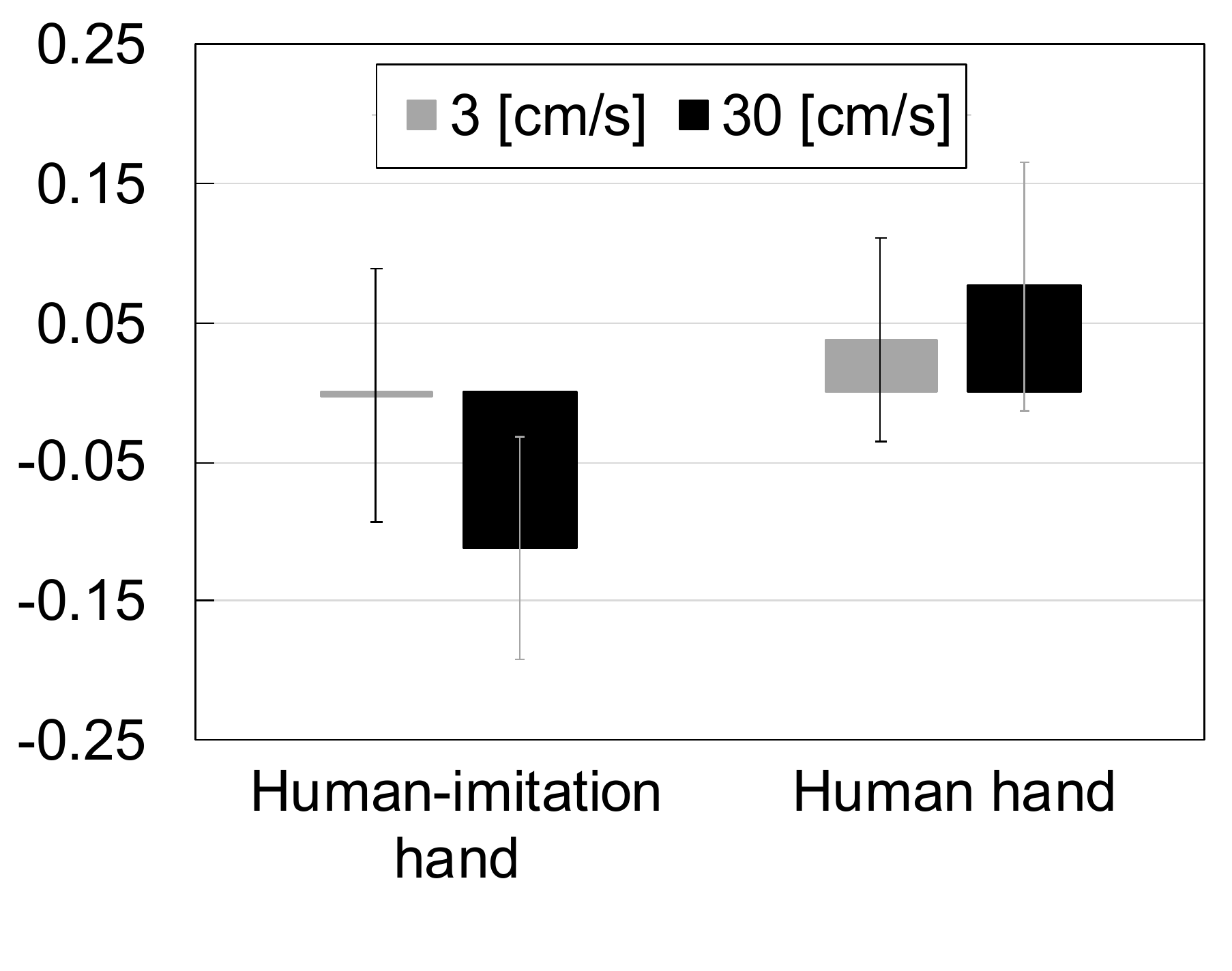}
          \hspace{1.6cm} B. Zyg
        \end{center}
      \end{minipage}
      
      % 1
      \begin{minipage}{0.3\hsize}
        \begin{center}
          \includegraphics[keepaspectratio=true, width=1.0\columnwidth]{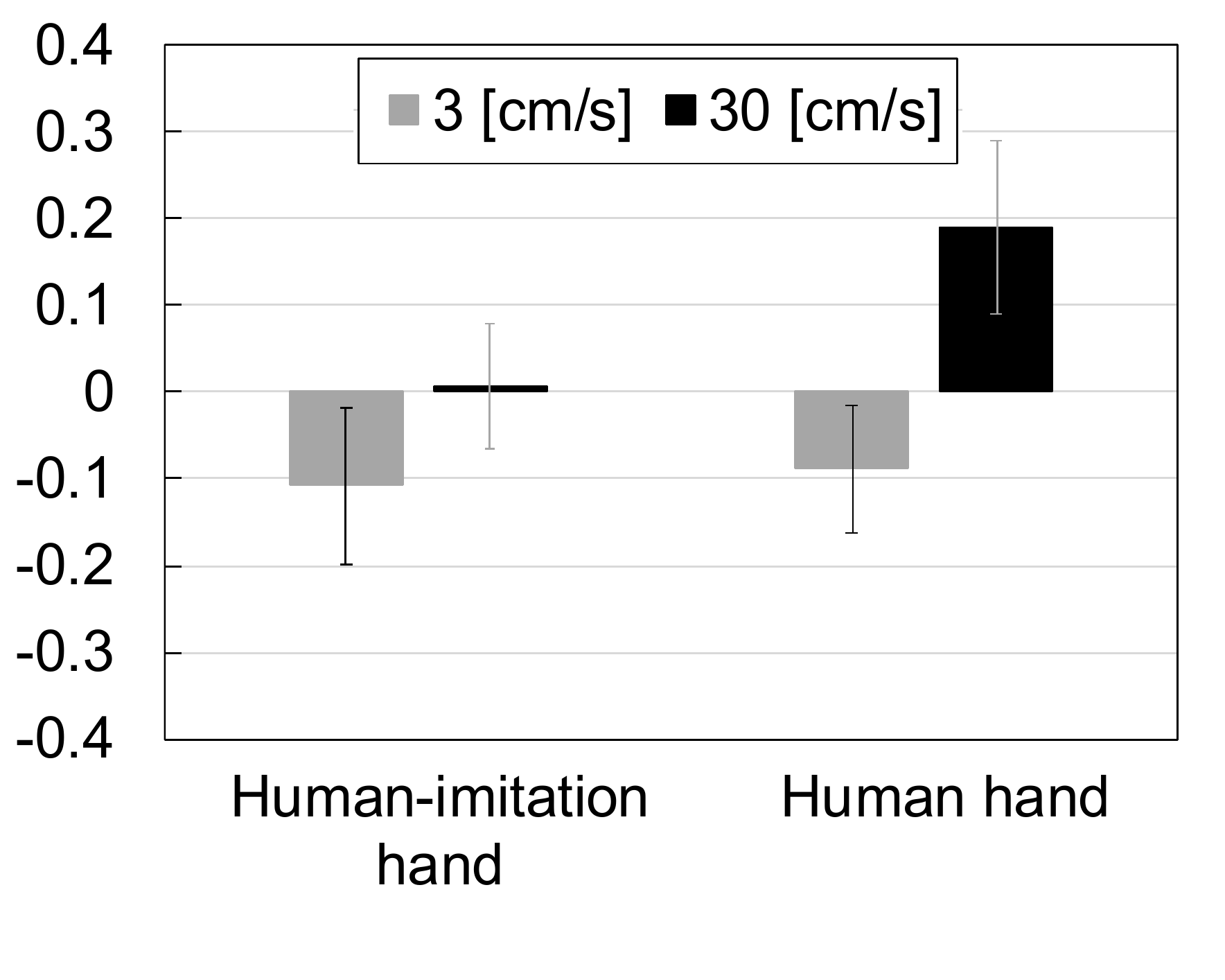}
         \hspace{1.6cm} C. SCL
        \end{center}
      \end{minipage}
    \end{tabular}
    \caption{The results of physiological evaluation.}
    \label{physiological}
  \end{center}
\end{figure*}
%%%%%%%%%%%%%%%%%%%%%%%%%%%%%%%%%%%%%%%%%%%%%%%%

\subsubsection{Free-description Type Questionnaire}
% To directly ask the different feelings of hand types and stroke speeds, we use a free-description type questionnaire regarding the human-likeness of the hands and stroke motion after the experiment.
We used a free-description type questionnaire to ask subjects about their different feelings regarding the human likeness of the hands and stroke motion after the experiment.
% Hence, we ask two questions that are “Please fill in freely about human likeness of the stroke motion and tactile feeling” and “Please fill in freely about the stroking motion and tactile feeling throughout the experiment.”
As such, we asked two questions: “Please fill in freely about human likeness of the stroke motion and tactile feeling” and “Please fill in freely about the stroking motion and tactile feeling throughout the experiment.”

\section{Results}
% Fig.~\ref{subjective} and Fig.~\ref{physiological} showed the experimental results of the subjective and physiological evaluations.
Figs.~\ref{subjective} and~\ref{physiological} show the experimental results of the subjective and physiological evaluations.
% As subjective evaluations, the higher rating of valence and arousal indicated that the subjects felt pleasant and sleepy.
From the subjective evaluations, the higher rating of valence and arousal shows that the subjects felt pleasant and sleepy.
% As physiological evaluations, the higher rating of Cor, Zyg, and SCL above zero indicated that the subjects felt unpleasant, pleasant, and arousal.
From the physiological evaluations, the higher rating of the Cor, Zyg, and SCL (above zero) shows that the subjects felt unpleasant, pleasant, and arousal.
% In the contrary, the lower the rating of Cor, Zyg, and SCL below zero indicated that the subjects felt pleasant, unpleasant, and sleepiness.
By contrast, a lower rating of the Cor, Zyg, and SCL (below zero) shows that the subjects felt pleasant, unpleasant, and sleepiness.

% There are both the dependent variables as subjective ratings (valence, arousal) or physiological ratings (Cor, Zyg, and, SCL) and the independent variables as two factors (agent, speed).
In this experiment, there are dependent variables such as subjective ratings (valence, arousal) or physiological ratings (Cor, Zyg, and SCL) and independent variables such as the two key factors studied (agent, speed).
% We used MANOVAs (Multivariate ANOVA)~\cite{tabachnick2007} to analyze differences within the independent variables to find out if the stroking motion of the robot evokes the same emotions as in humans.
We used MANOVAs (multivariate ANOVA) to analyze the differences within the independent variables to determine whether the stroking motion of the robot evoked the same emotions as with human touch.
%%%%%%%%%%%%
% The MANOVA for the subjective ratings showed significant main effects of hand and speed ({\em Wilks'}$\lambda F(2,10) = 9.71$ and $4.74$, $ps < 0.05$) with a non-significant trend of interaction.
% In physiological responses, only the main effect of speed was signiﬁcant ({\em Wilks'}$\lambda$ $F(3,9) = 4.72$, $p < 0.05$).
% Follow-up univariate ANOVAs 
% were conducted for significant effects. 
% Table~\ref{subjectiveANOVA} and \ref{physiologicalANOVA} show the results of ANOVA.
% The main effect of the agent was significant in valence. The main effect of the speed was significant in the arousal and the Cor.
%%%%%%%%%%%%
The MANOVA for the subjective ratings showed significant main effects of hand and speed ({\em Wilks'}$\lambda F(2,10) = 9.71$ and $4.74$, $ps < 0.05$) with a non-significant trend of interaction.
For physiological responses, only the main effect of speed was significant ({\em Wilks'}$\lambda$ $F(3,9) = 4.72$, $p < 0.05$).
Follow-up univariate simple-effect contrasts (two-tailed) were performed for significant effects. 
For the valence ratings, only the effect of hand was significant ($p < 0.001$), indicating that the human hand is more pleasant than the human-imitation hand.
For the arousal ratings, only the effect of speed was significant ($p < 0.05$), indicating higher arousal in response to 30 cm/s touch than to 3 cm/s touch.
For the Cor activity, the effect of speed was significant ($p < 0.05$), indicating higher Cor activity under the 3 cm/s than 30 cm/s condition.
For the SCL, the effect of speed was significant ($p < 0.05$), indicating higher responses to 30 cm/s than 3 cm/s touch.
Thus, we obtained three specific significant differences.
\begin{itemize}
\item The human hand was more pleasant than the human-imitation hand on valence.
\item The 3 cm/s speed was sleepier than the 30 cm/s speed on arousal.
\item The 30 cm/s speed was more pleasant than the 3 cm/s speed on valence.
\end{itemize}

\section{Discussion}
\subsection{Slow Stroking vs. Rapid Stroking}
% The slow stroking caused sleepiness and low pleasantness.
Slow stroking caused sleepiness and low pleasantness. 
% Therefore, question $1$ was answered that subjects felt more pleasant and arousal by 30 cm/s speed than 3 cm/s speed in both the human and the robot.
Therefore, the answer to Question 1 suggests that subjects experienced more pleasant and arousal feelings from the 30 cm/s speed than the 3 cm/s speed with both the human and the robot. 
% This tendency is the same even if the agent is a robot or a human.
This tendency was the same whether the agent was a robot or a human. 
% Though the results of valence were different from the finding of the related works~\cite{loken2009coding, pawling2017c}, we thought it were caused that the different number of skin receptors between the arm and the back.
Although the results of valence were different from the findings of related studies~\cite{loken2009coding, pawling2017c}, we believe it to be caused by the different number of skin receptors between the arm and the back.

% Slow stoke is preferred in tactile care~\cite{tactilemassage}.
The slow stoke is preferred in tactile care~\cite{tactilemassage}. 
% Slow stroke caused low pleasant, but provides sleepiness, in other words, calm.
The slow stroke causes low pleasantness, but sleepiness, in other words, a sense of calm.
% In a caregiving situation, a patient's mind would like to be kept calm, not get arousal.
In a caregiving situation, a patient’s mind should be kept calm and not aroused. 
% We think that is the reason for the preference for slow stroke.
We believe that this is the reason for the slow-stroke preference. 

% Note that a few subjects were more pleasant by 3 cm/s speed than 30 cm/s speed.
% By comparing in the pleasantness score size within the subject that four subjects felt more pleasant to stroke 3 cm/s speed than 30 cm/s speed.
% Therefore, note that there were a little individual differences in the speed.

% For example, if we provide pleasantness to a subject by the robot, we choose that the speed of the robot stoke is 30 cm/s.
% In addition, if we provide sleepiness to a subject by the robot, we choose that the speed of the robot stoke is 3 cm/s.
% Therefore, we consider that it is necessary to stroke slowly to make relax for a subject even if it is a little unpleasant.

\subsection{Human-Imitation Hand vs. Human Hand}
By comparing the human-imitation hand with the human hand, we confirmed from the subjective evaluation that the human-imitation hand exhibited slightly lower pleasantness than the human hand.
% But from the questionnaire, we found the room to improve the hand by keeping the large touching area.
However, from the questionnaire, we found room to improve the hand by keeping a large touch area.
% In the questionnaire, two subjects commented that ``I felt more human-like when a touching area is larger.''
In the questionnaire, two subjects commented that “It felt more human-like when the touching area is larger.”
% In addition, two subjects commented that ``The hand didn't fit closely when my back was bent'' and ``Sometimes I felt that the robot stroke provided me with along my back, and sometimes it did not provide.''
In addition, two subjects commented that “The hand did not fit closely when my back was bent” and “Sometimes I felt that the robot stroke ran along my back, and sometimes it did not.”
% Note that this subject may think a change of the touching area is one of the conditions, not caused by the difference of the agent.
Note that this subject may have thought that a change in the touching area was one of the conditions and not something caused by the different agents.

% We suggested the reason to support the similar emotions from the free-description type questionnaires that six subjects' comments were ``The warm temperature provides a feeling of human-likeness'' and ``The robot hand fitted much on the back.''
% Thus, we suggest that similar emotions are evoked because of the flexibility, softness, and warmness of the human-imitation hand.

% We believe that the concept of the design of the human-imitation hand is correct, from the positive comment of the questionnaires.
We believe that the concept of the design of the human-imitation hand is correct, based on the positive comments from the questionnaires.
% For example, six subjects’ comments were “The warm temperature provides a feeling of human-likeness” and “The robot hand fitted the back much.”
For example, six subjects’ comments were “The warm temperature provides a feeling of human-likeness” and “The robot hand fit my back well.” 
% Realizing a large touching area is our tackling issue, which is solved by a fitting mechanism and more flexible control of the arm.
Realizing a large touching area is one area for us to address, which can be solved using a fitting mechanism and a more flexible arm control.

% However, we considered two causes for the difference between agents by the questionnaire results. 
% First, the shape of the human-imitation hand was not suitable for a large contacting surface as the CM joint being forward.
% Second, the robot motion did not completely fit the arched back because stroke motion was simple straight-line.
% From these results, the proposed system has a room to improve; we should consider the mechanism and control to fit the back more.

\section{Conclusions}
% In this paper, we proposed a human-imitation hand and the robot system that gently strokes the back of a human to provide us a positive emotional effect.
In this paper, we proposed a human-imitation hand and a robot system that gently strokes the back of a human to provide a positive emotional effect.
% For this purpose, we developed the human-imitation hand which was endowed with human-like flexibility, softness, and warmness.
For this purpose, we developed a human-imitation hand that was endowed with human-like flexibility, softness, and warmth.
% And we compared the system with human stroke motions.
We also compared the system with human stroke motions.
% In detail, we evaluated the effect for 1) the human hand and the human-imitation hands and 2) the stroking speed (3 and 30 cm/s) with both subjective and physiological measures. 
With both subjective and physiological measures, we evaluated the effects of:
\begin{enumerate}
\renewcommand{\labelenumi}{\arabic{enumi})}
 \item the human and human-imitation hands;
 \item the stroking speed (3 and 30 cm/s).
\end{enumerate}

% From the experiment results, we reached the following two conclusions.
% First, the subjects evaluated strokes at a similar tendency in the speed aspect independent of the human-imitation hand and human hand. Both subjective and physiological evaluations support this evaluation. Second, the subjects seem to feel more pleasant with faster-speed (30 cm/s than 3 cm/s) stroke.
% However, we confirmed a difference of contact area and stroke motion between the robot and the human in terms of pleasantness, and the proposed system has a room to improve.
% In particular, we should consider the mechanism and control to fit the back more.
% Though affective touch by a robot is a little inferior to affective touch by a human, we find out three interest findings. 
Though affective touch using a robot is a little inferior to affective touch by a human, we find three interesting findings. 
% First, the subjects evaluate strokes at a similar tendency in the speed aspect with both human hand and human-imitation hand in both subjective and physiological evaluations. 
First, the subjects evaluated strokes similarly with regard to the stroke speed of the human and human-imitation hand, in both the subjective and physiological evaluations. 
% Second, the subjects feel more pleasant and arousal with the faster-speed (30 cm/s than 3 cm/s) stroke.
Second, the subjects felt more pleasantness and arousal with a faster stroke (30 cm/s rather than 3 cm/s). 
% Third, less fitting due to bending a back gets the bad effect and there is the room to enhance the system by improving the fitting mechanism of the proposed hand.
Third, poorer fitting of the human-imitation hand due to the bending of the back had a negative effect on the subjects.
We found room to enhance the system by improving the fitting mechanism of the proposed hand.

% In future work, we will improve the human-imitation hand to fit the arched back of a subject more.
In future work, we aim to improve the human-imitation hand to fit the arched back of a subject.
% And further studies are needed to evoke the most pleasant by comparing various speeds.
Further studies are needed to evoke the most pleasantness by comparing various speeds.
% Furthermore, we would like to investigate the emotional effect of the difference in ages by recruiting elderly subjects.
Furthermore, we aim to investigate the emotional effect of the difference in age by recruiting elderly subjects.

\section*{Acknowledgment}
This work was supported by JSPS KAKENHI Grant Num-
ber JP17K13088 and JP19H01124, JST Research Complex Promotion Program, and JST CREST Grant Number JPMJCR17A5, Japan.
%This work was supported by JST CREST Grant Number JPMJCR17A5, Japan.

\bibliographystyle{IEEEtran}
\footnotesize
\bibliography{bibliography/reference}

\begin{thebibliography}{10}
\providecommand{\url}[1]{#1}
\csname url@rmstyle\endcsname
\providecommand{\newblock}{\relax}
\providecommand{\bibinfo}[2]{#2}
\providecommand\BIBentrySTDinterwordspacing{\spaceskip=0pt\relax}
\providecommand\BIBentryALTinterwordstretchfactor{4}
\providecommand\BIBentryALTinterwordspacing{\spaceskip=\fontdimen2\font plus
\BIBentryALTinterwordstretchfactor\fontdimen3\font minus
  \fontdimen4\font\relax}
\providecommand\BIBforeignlanguage[2]{{%
\expandafter\ifx\csname l@#1\endcsname\relax
\typeout{** WARNING: IEEEtran.bst: No hyphenation pattern has been}%
\typeout{** loaded for the language `#1'. Using the pattern for}%
\typeout{** the default language instead.}%
\else
\language=\csname l@#1\endcsname
\fi
#2}}

\bibitem{henricson2008outcome}
M.~Henricson, A.~Ersson, S.~M{\"a}{\"a}tt{\"a}, K.~Segesten, and A.-L.
  Berglund, ``The outcome of tactile touch on stress parameters in intensive
  care: a randomized controlled trial,'' \emph{Complementary therapies in
  clinical practice}, vol.~14, no.~4, pp. 244--254, 2008.

\bibitem{andersson2009tactile}
K.~Andersson, L.~T{\"o}rnkvist, and P.~W{\"a}ndell, ``Tactile massage within
  the primary health care setting,'' \emph{Complementary therapies in clinical
  practice}, vol.~15, no.~3, pp. 158--160, 2009.

\bibitem{suzuki2010physical}
M.~Suzuki, A.~Tatsumi, T.~Otsuka, K.~Kikuchi, A.~Mizuta, K.~Makino, A.~Kimoto,
  K.~Fujiwara, T.~Abe, T.~Nakagomi, \emph{et~al.}, ``Physical and psychological
  effects of 6-week tactile massage on elderly patients with severe dementia,''
  \emph{American Journal of Alzheimer's Disease \& Other
  Dementias{\textregistered}}, vol.~25, no.~8, pp. 680--686, 2010.

\bibitem{toyoshima2018WELCARO}
K.~Toyoshima, M.~Ding, J.~Takamatsu, and T.~Ogasawara, ``What is required for a
  robot to gently stroke a human using its hand,'' in \emph{Proc. of the ICRA
  2018 Workshop on Elderly Care Robotics–Technology and Ethics}, 2018.

\bibitem{koizumi2017}
Y.~Koizumi, Y.~Kohno, Y.~Matsui, and K.~Sakai, ``Physiological, biochemical and
  psychological verification of the relaxation effects of touch care on touch
  care practitioners,'' \emph{Journal of Nursing Science and Engineering},
  vol.~4, no.~1, pp. 27--38, 2017.

\bibitem{loken2009coding}
L.~S. L{\"o}ken, J.~Wessberg, I.~Morrison, F.~McGlone, and H.~Olausson,
  ``Coding of pleasant touch by unmyelinated afferents in humans,''
  \emph{Nature Neuroscience}, vol.~12, no.~5, pp. 547--548, 2009.

\bibitem{wada2004paro}
K.~Wada, T.~Shibata, T.~Saito, and K.~Tanie, ``Effects of robot-assisted
  activity for elderly people and nurses at a day service center,'' in
  \emph{Proc. of the IEEE}, vol.~92, no.~11, 2004, pp. 1780--1788.

\bibitem{ogawa2011telenoid}
K.~Ogawa, S.~Nishio, K.~Koda, G.~Balistreri, T.~Watanabe, and H.~Ishiguro,
  ``Exploring the natural reaction of young and aged person with telenoid in a
  real world,'' \emph{Journal of Advanced Computational Intelligence and
  Intelligent Informatics}, vol.~15, no.~5, pp. 592--597, 2011.

\bibitem{king2010}
C.-H. King, T.~L. Chen, A.~Jain, and C.~C. Kemp, ``Towards an assistive robot
  that autonomously performs bed baths for patient hygiene,'' in \emph{Proc. of
  the 2010 IEEE International Conference on Intelligent Robots and Systems
  (IROS)}, 2010, pp. 319--324.

\bibitem{chen2011}
T.~L. Chen, C.-H. King, A.~L. Thomaz, and C.~C. Kemp, ``Touched by a robot: An
  investigation of subjective responses to robot-initiated touch,'' in
  \emph{Proc. of the 6th ACM/IEEE International Conference on Human-Robot
  Interaction (HRI)}, 2011, pp. 457--464.

\bibitem{humanitude}
Y.~Gineste and J.~Pellissier, \emph{Humanitude: comprendre la vieillesse,
  prendre soin des hommes vieux}.\hskip 1em plus 0.5em minus 0.4em\relax Armand
  Colin, 2007.

\bibitem{honda2019HAI}
S.~Honda, T.~Sawabe, S.~Nishimura, W.~Sato, Y.~Fujimoto, A.~Plopski,
  M.~Kanbara, and H.~Kato, ``Evaluation of relationship between stroke pace and
  speech rate for touch-care robot,'' in \emph{Proc. of the 7th International
  Conference on Human-Agent Interaction (HAI)}, 2019, pp. 283--285.

\bibitem{nakanishi2014}
H.~Nakanishi, K.~Tanaka, and Y.~Wada, ``Remote handshaking: touch enhances
  video-mediated social telepresence,'' in \emph{Proc. of the SIGCHI Conference
  on Human Factors in Computing Systems (CHI '14)}, 2014, pp. 2143--2152.

\bibitem{cabibihan2015}
J.-J. Cabibihan, D.~Joshi, Y.~M. Srinivasa, M.~A. Chan, and A.~Muruganantham,
  ``Illusory sense of human touch from a warm and soft artificial hand,''
  \emph{IEEE Transactions on Neural Systems and Rehabilitation Engineering},
  vol.~23, no.~3, pp. 517--527, 2015.

\bibitem{ueno2019ROBIO}
A.~Ueno, I.~Mizuuchi, and Y.~Morooka, ``Proposal for robot hand and forearm
  design to reproduce human-to-human physical contact,'' in \emph{Proc. of the
  2019 IEEE International Conference on Robotics and Biomimetics (ROBIO)},
  2019, pp. 971--976.

\bibitem{ueno2020ROMAN}
A.~Ueno, V.~Hlav{\'{a}}{\v{c}}, I.~Mizuuchi, and M.~Hoffmann, ``Touching a
  human or a robot? investigating human-likeness of a soft warm artificial
  hand,'' in \emph{Proc. of the 29 IEEE International Symposium on Robot and
  Human Interactive Communication (RO-MAN).}, 2020.

\bibitem{pawling2017c}
R.~Pawling, P.~R. Cannon, F.~P. McGlone, and S.~C. Walker, ``C-tactile afferent
  stimulating touch carries a positive affective value,'' \emph{PloS one},
  vol.~12, no. 3, e0173457, 2017.

\bibitem{weinstein1968intensive}
S.~Weinstein, ``Intensive and extensive aspects of tactile sensitivity as a
  function of body part, sex and laterality,'' \emph{The skin senses}, pp.
  195--222, 1968.

\bibitem{tactilemassage}
W.~E. Arnould-Taylor, \emph{The Principle and Practice of Physical Therapy},
  3rd~ed.\hskip 1em plus 0.5em minus 0.4em\relax Cheltenham: Stanley Thomas,
  1991.

\bibitem{russell1989affect}
J.~A. Russell, A.~Weiss, and G.~A. Mendelsohn, ``Affect grid: a single-item
  scale of pleasure and arousal.'' \emph{Journal of personality and social
  psychology}, vol.~57, no.~3, pp. 493--502, 1989.

\bibitem{mayo2018putting}
L.~M. Mayo, J.~Lind{\'e}, H.~Olausson, M.~Heilig, and I.~Morrison, ``Putting a
  good face on touch: Facial expression reflects the affective valence of
  caress-like touch across modalities,'' \emph{Biological psychology}, vol.
  137, pp. 83--90, 2018.

\bibitem{ree2019touch}
A.~Ree, L.~M. Mayo, S.~Leknes, and U.~Sailer, ``Touch targeting c-tactile
  afferent fibers has a unique physiological pattern: A combined electrodermal
  and facial electromyography study,'' \emph{Biological psychology}, vol. 140,
  pp. 55--63, 2019.

\bibitem{fridlund1986guidelines}
A.~J. Fridlund and J.~T. Cacioppo, ``Guidelines for human electromyographic
  research,'' \emph{Psychophysiology}, vol.~23, no.~5, pp. 567--589, 1986.

\bibitem{sato2020facial}
W.~Sato, K.~Minemoto, A.~Ikegami, M.~Nakauma, T.~Funami, and T.~Fushiki,
  ``Facial emg correlates of subjective hedonic responses during food
  consumption,'' \emph{Nutrients}, vol.~12, no.~4, p. 1174, 2020.

\end{thebibliography}
\end{document}